\documentclass[letterpaper, 10 pt, conference]{ieeeconf}  

\IEEEoverridecommandlockouts                              

\makeatletter
\let\NAT@parse\undefined
\makeatother
\usepackage[numbers]{natbib}

\usepackage[pdftex]{graphicx}
\usepackage{amsmath} 
\usepackage{amssymb}  
\usepackage{subfigure}
\usepackage{multirow}

\usepackage{irap_SIunits}
\usepackage{irap_acronyms}
\usepackage{irap_math}
\usepackage{irap_misc}
\usepackage{hyperref}

\usepackage{soul,color}

\usepackage{xcolor}

\title{\LARGE \bf
Generic Camera Attribute Control using Bayesian Optimization
}

\author{Joowan Kim${}^{1}$, Younggun Cho${}^{1}$ and Ayoung Kim${}^{1*}$
\thanks{J. Kim, Y. Cho, and A. Kim are with the Department of Civil and Environmental Engineering,
        KAIST, Daejeon, S. Korea \texttt{[jw\_kim, yg.cho, ayoungk]@kaist.ac.kr}
        }%
\thanks{This material is based upon work supported by KI Robotics and
        by Korea MOLIT via `U-City Master and Doctor Course Grant Program'.}
}

\begin{document}

\maketitle
\thispagestyle{empty}
\pagestyle{empty}

\begin{abstract}

Cameras are the most widely exploited sensor in both robotics and computer
vision communities. Despite their popularity, two dominant attributes (i.e.,
gain and exposure time) have been determined empirically and images are captured
in very passive manner. In this paper, we present an active and generic camera
attribute control scheme using Bayesian optimization. We extend from our
previous work \cite{jkim-2018-icra} in two aspects. First, we propose a method
that jointly controls camera gain and exposure time. Secondly, to speed up the
Bayesian optimization process, we introduce image synthesis using the \ac{CRF}.
These synthesized images allowed us to diminish the image acquisition time
during the Bayesian optimization phase, substantially improving overall control
performance.  The proposed method is validated both in an indoor and an outdoor environment where light condition rapidly changes. Supplementary material is available at \url{https://youtu.be/XTYR_Mih3OU}.

\end{abstract}

\section{Introduction}
\label{sec:intro}

Despite the wide popularity of cameras in many vision-based applications, visual
perception possesses a critical limitation when changing light conditions alter
appearance. Many studies have focused on the critical error caused by this
\ac{HDR} environment by proposing an algorithmic compensation to incorporate
varying illumination. Aside from these algorithmic efforts, some studies have
focused on a hardware solution. Early studies introduced camera attribute
control for adequate adjustment of the camera hardware.


The three dominant factors determining image quality are aperture,  exposure
time, and gain. Aperture is often adjusted manually and determines the amount of
incoming light. This amount of incoming light is constant once the aperture is
fixed. The next factor, exposure time, is controlled by the camera's shutter
speed. Early studies have presented exposure adjustment via shutter speed
control as in \cite{shim2014auto} and \cite{zhang2017active}. Proper exposure
time is required because a longer exposure time may result in the frame rate to
drop and the image to blur. Lastly,  gain controls the signal amplification of
the sensor. The higher the gain, the brighter an image. However, because gain
amplifies all the signals in the image, the noise components are also amplified.

\begin{figure}[!h]
  \centering%
    \includegraphics[width=0.95 \columnwidth]{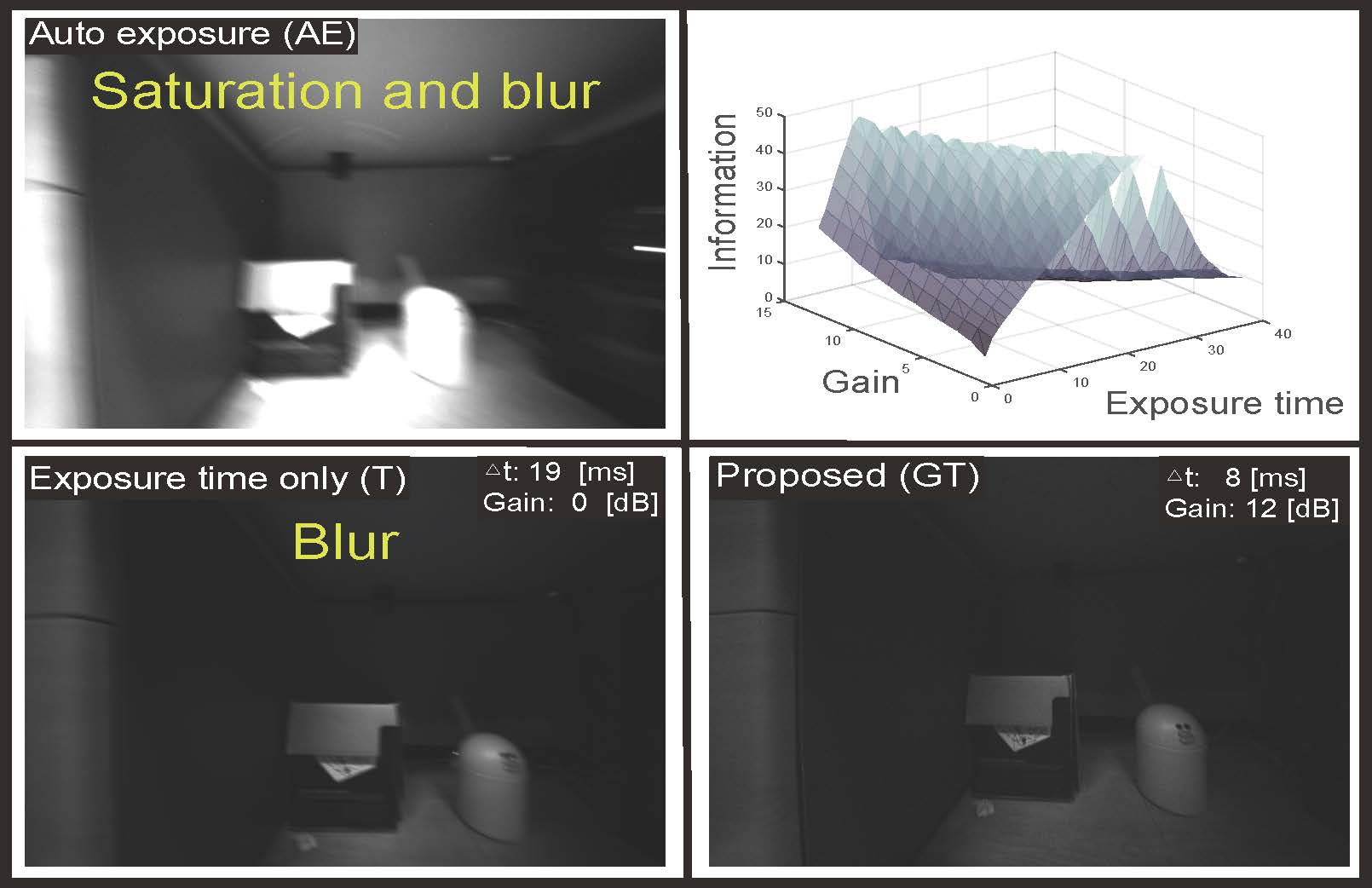}

  \caption{Captured image comparison among auto exposure (\texttt{AE}), exposure
  time only (\texttt{T}), and proposed generic attribute control (\texttt{GT}).
  The optimal image contains more information (i.e., magnitude of the gradient)
  little saturation, blurring, and noise. Left column images illustrate the
  problem of separately controlling exposure time and gain. Independent control
  of them in a dark environment can result in image blurring caused by exposure
  and noise due to gain. By applying the proposed method, however, these two
  parameters are controlled simultaneously (bottom right) producing the improved
  image.}

  \label{fig:intro}
\end{figure}

In conventional approaches, the two attributes are considered in a passive
manner. For exposure time selection, many vision-based approaches rely on either
automatic exposure control built into the camera or a fixed exposure value
assuming a constant brightness in an environment. For the choice of gain, the
increase of the gain was prohibited to avoid noisy images. But most critically,
the two attributes were tackled separately.

\begin{figure*}[!t]
  \centering%
  \includegraphics[width=0.9\textwidth, ]{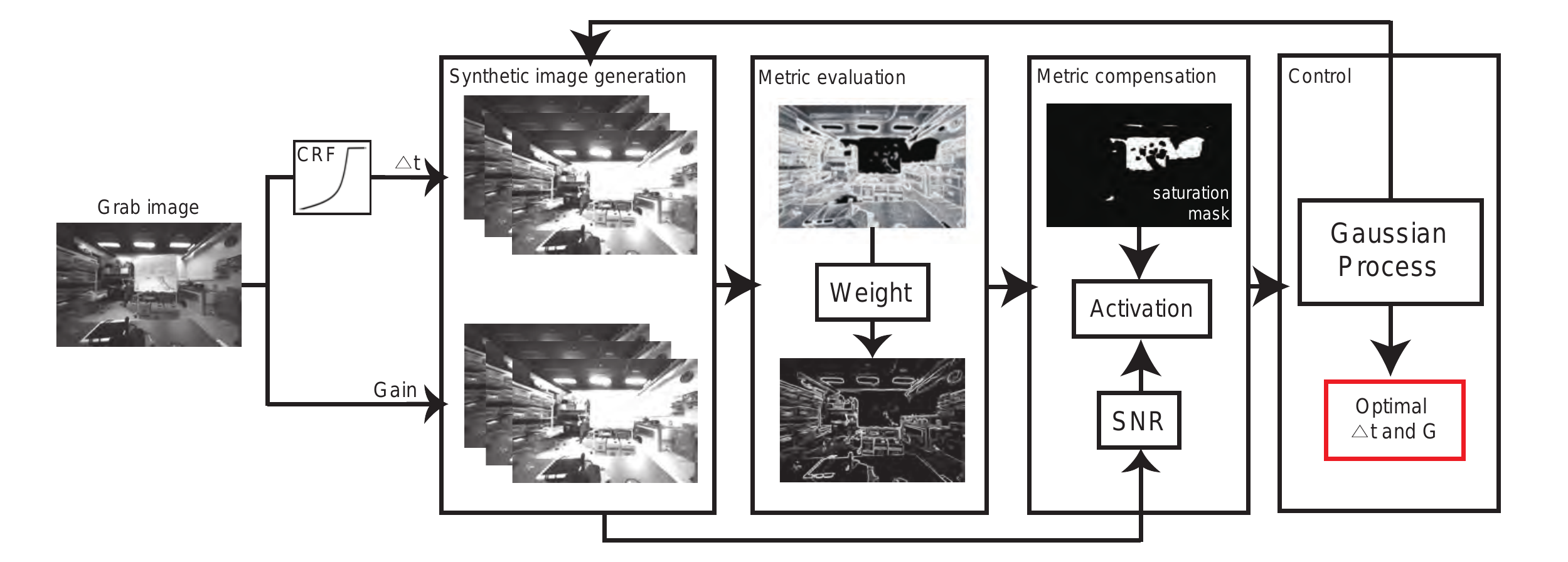}

  \caption{Diagram for the proposed method. The overall procedure consists of
  four modules, (\textit{i}) image synthesis, (\textit{ii}) metric evaluation,
  (\textit{iii}) metric compensation and (\textit{iv}) control module.}

  \label{fig:overview}
\end{figure*}

Simultaneous control of these two attributes is beneficial in many aspects. When
the camera undergoes extreme light condition, careful gain control is essential.
For a dark scene, meaningful camera control is possible only when the gain
partakes.  Our finding is that the two parameters are closely related as in
\figref{fig:intro} and an unified controller needs to solve for the optimal
attribute in a generic manner.


In this paper, we extend our previous exposure control by associating it with
camera gain control. The proposed method is novel presenting the following
contributions:


\begin{itemize}

\item We present a generic approach for active camera attribute control by
considering both exposure time and gain. While these two factors are often
determined separately, we report a thorough experiments to analyze the
associative effect of exposure time and gain under a light varying environment.

\item The main limitation of our previous work \cite{jkim-2018-icra} was at the
cost of evaluating a function by capturing  real images with a specific exposure
time. We improve the acquisition module by introducing image synthesis. As a
result, we achieved a substantial improvement in attribute control for
performance.



\end{itemize}

\section{Related Works}
\label{sec:related}

Among the camera attribute capturing images, gain and exposure time controls
were widely studied. Exposure control has been examined especially in the
\ac{HDR} environment. Early works on the \ac{HDR} environment mostly intended
to compensate the pixel loss from saturation by using a \ac{CRF}

\textbf{\textit{Image Fusion in \acs{HDR}}} \citeauthor{debevec1997recovering}
\cite{debevec1997recovering} defined the relationship between the \ac{CRF} and
radiance value through linear mapping and created a synthetically computed
radiance map to reconstruct an \ac{HDR} image that is suitable for human vision.
In \cite{lin2005determining}, the authors introduced the camera's radiometric
response function to a single grayscale image without using a registered set of
images. Using the statistical properties of the grayscale histogram in the edge
area, the authors obtained information about the radiometric calibration.


\textbf{\textit{Exposure time control}} In recent years, research has been
conducted on finding appropriate exposure times in the robotics field
\cite{shim2014auto}, \cite{zhang2017active}, and \cite{lu2010camera}.
Shim~\cite{shim2014auto} proposed using a method that selects the exposure time
of a camera by examining the largest gradient magnitude of an image. The
synthetic images were generated using a gamma function and mapped to the
exposure time.  Zhang~\cite{zhang2017active} used the gradient percentile to
express the image information amount by weighting the gradient. They also
examined a photometric response function to control the exposure time. Another
method~\cite{lu2010camera} is to leverage image entropy to measure the amount of
information in an image and adjust the exposure time and gain accordingly. By
doing so, the image parameter with the highest entropy is selected while no
control scheme was considered.

\textbf{\textit{Gain control}} \ac{AGC} is a camera function that increases the
average gain when the image is too dark and reduces it when the image is too
bright \cite{fowler2004automatic}. In order to prevent the gain from
oscillating, the increase and decrease of the gain between adjacent frames is
limited to one. \citeauthor{litvinov2005addressing}
\cite{litvinov2005addressing} proposed a solution that utilizes a radiometric
response function to simultaneously estimate the unknown response function and
camera gain (exposure) in the image sequences.


\section{Generic Camera Attribute Control}
\label{sec:method}



We now introduce a generic camera attribute controller that simultaneously
controls both gain and exposure time. The overall process is as shown in
\figref{fig:overview}. Results from an improper assignment of these two
attributes are critical. The improper exposure time may blur and saturate the
resulting images; change in gain produces additional noise when a higher gain is
applied.  To properly handle these factors, we modified our previous image
quality evaluation metric to include the noise caused by gain control.


\subsection{Image Information Measure using SNR}
\label{sec:info}

Let us start with the image quality evaluation metric. We variate from the
metric introduced in our previous work \cite{jkim-2018-icra} to include not only
saturation from exposure but also noise from an increase in gain. For an image
$I$, the proposed evaluation metric $G_I$ is the summation of metric from the
exposure time $(G_t)_I$ and metric from the gain $(G_k)_I$
\begin{equation}
  G_I = (G_t)_I - \frac{1}{\kappa} (G_k)_I
  \label{G_all}
\end{equation}
, where $\kappa$ is the user parameter controlling the penalization balance
between the exposure time and the gain. In this work $\kappa=5$ was used. The
computed metric is then compensated using a saturation mask and activation
function. We adopted our previous module for this compensation
\cite{jkim-2018-icra}.

The first term, $(G_t)_I$, is the \ac{EWG} from our previous work
\cite{jkim-2018-icra}. For a pixel $i$, we compute $(g_t)_i$ using magnitude of gradient
$\parallel \nabla I(i) \parallel^2$, entropy $H_I$, activation function
$\pi(\cdot)$, image mask $M_i$, and entropy weight $W_i$ as below.
\begin{equation}
  \scriptsize
  (g_t)_i = W_i \parallel \nabla I(i) \parallel^2 + \pi(H_i) M_i(H_i) W_i \frac{1}{N}\sum_{j=0}^{N-1} \parallel \nabla I(j) \parallel^2
 \label{gt_i}
\end{equation}
The overall \ac{EWG} for entire image $I$ is the summation of $g_i$ over $N$
pixels $(G_t)_I = \sum_{i}^N (g_t)_i$. This \ac{EWG} consists of a term for
image gradient and a term for saturation penalty. We refer to our previous
derivation in \cite{jkim-2018-icra} for detail of each term.

When taking gain into consideration, we need to account for the subsequent
increase in noise. To incorporate the increase in noise, we adopted a \ac{SNR}
as a physical measure of sensitivity to noise. The \ac{SNR} of an image is
typically calculated as the ratio of the mean pixel to the standard deviation of
the pixel for a given neighbor. In general, the lower the gain, the higher the
\ac{SNR} and the better the image quality. We propose including this \ac{SNR} in
the image evaluation metric to penalize the increase of noise by the gain.
Industry standards measure and define sensitivity in terms of the ISO film;
\unit{32}{dB} is defined as an excellent image quality and \unit{20}{dB} as an
acceptable image \cite{yang1999comparative}. In this paper, we weight the image
gradient using an \ac{SNR} ratio with respect to this acceptable $SNR_{ref} =
20$. We compute $(g_k)_i$ for each pixel $i$ as
\begin{equation}
  \small
  (g_k)_i = \left( 1 - \frac{SNR}{SNR_{ref}} \right) \left( \frac{1}{N}\sum_{j=0}^{N-1} \parallel \nabla I(j) \parallel^2 \right).
  \label{gk_i}
\end{equation}
Then, the \ac{SNR} metric for entire image $I$ is the summation of $(g_k)_i$
over $N$ pixels $(G_k)_I = \sum_{i}^N (g_k)_i$.

Given two terms for exposure time $(G_t)_I$ and gain $(G_k)_I$, we sum two terms
as in \eqref{G_all}.  We will denote this final metric, $G_I$, as \ac{NEWG}
throughout the paper.

\subsection{Camera Attribute Control}
\label{sec:ctrl}

\begin{figure}[!t]
	\centering%
	\def\width{0.45\columnwidth}%
	\subfigure[True distribution]{
		\includegraphics[trim = 40 20 35 35, clip, width=\width] {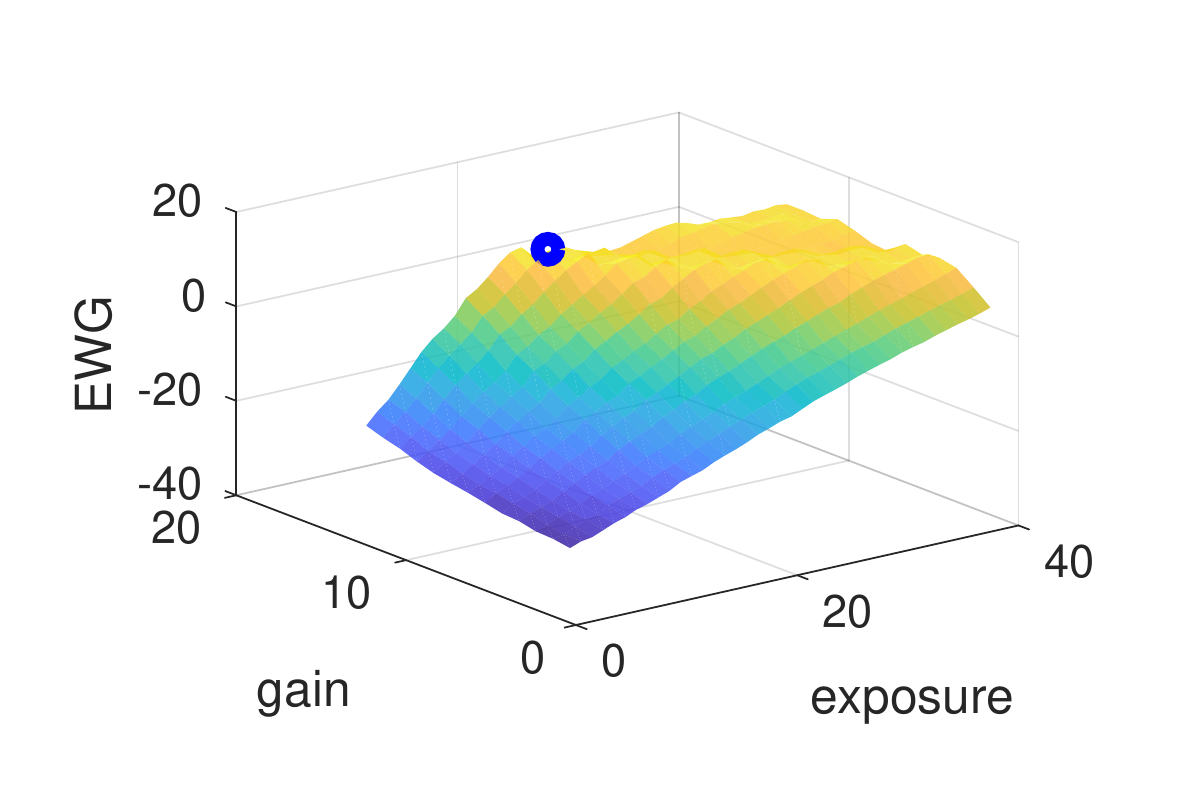} 
		\label{fig:ctrl_true}
	}
	\subfigure[Estimated distribution]{
		\includegraphics[trim = 40 20 35 35, clip, width=\width] {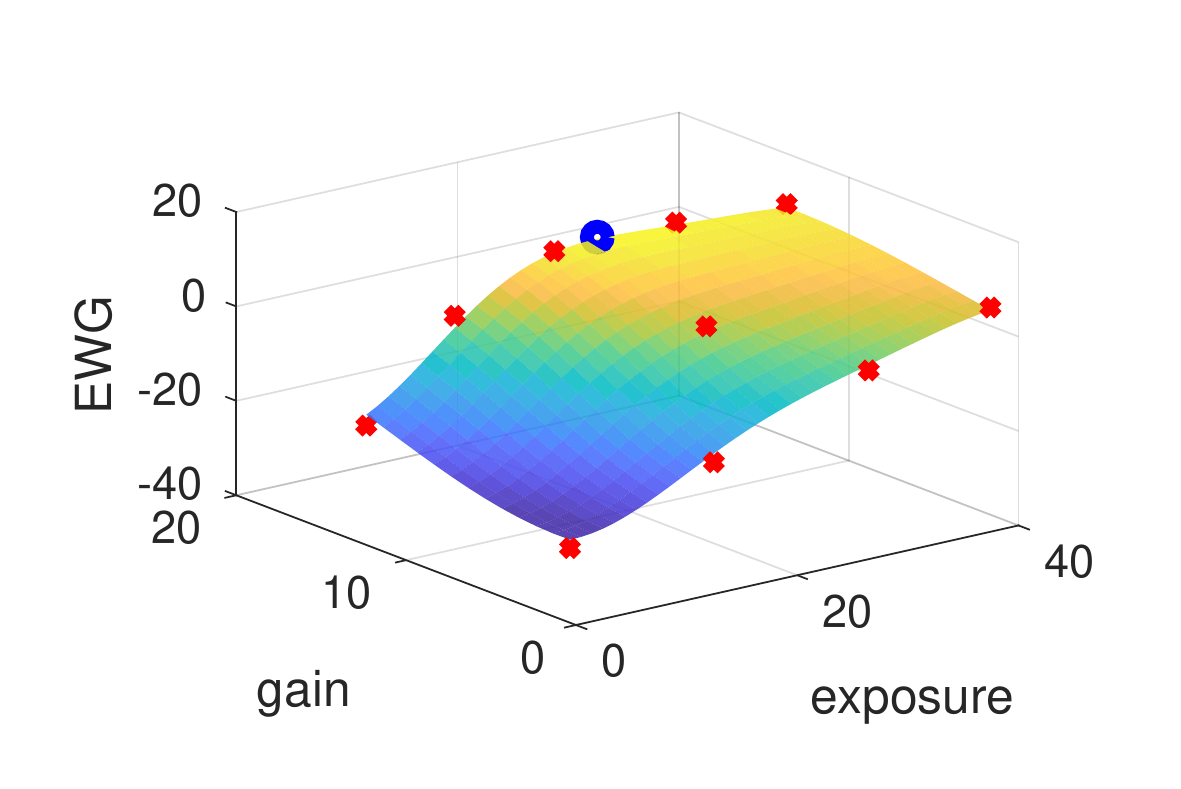} 
		\label{fig:ctrl_estim}
	}
	\caption{Example of camera attribute control. \subref{fig:ctrl_true} True distribution of \ac{NEWG} and true optimal values (blue circle). \subref{fig:ctrl_estim} Estimated distribution using training points (red x) and predicted optimal values (blue circle). }
	\label{fig:ctrl_fig}
\end{figure}

Using a metric that accounts for both exposure time and gain, we perform
attribute control by using Bayesian optimization. Because \ac{NEWG} analyzes
pixel distribution, which requires high levels of computation, we apply a
\ac{GP}-based optimization strategy to predict the best parameter for each
attribute. Our control module, an extension of our previous work
\cite{jkim-2018-icra}, jointly estimates the best exposure and gain values for
next image input.

%

We select stationary and simple kernel function, \ac{SE}. For our application,
the query exposure and gain spaces are bound by minimum and maximum exposure and
gain values. After restricting the input space of our control parameters, we
estimate optimal and fixed hyper-parameters via a log-likelihood optimization
technique. To derive varying hyper-parameters, we construct various training
datasets (exposure, gain and relative metrics) on several environments.


For camera attribute control, we use incremental learning with a maximum
variance acquisition function \cite{jkim-2018-icra}. Until termination
conditions, the camera's parameters are iteratively estimated as an exploration
task. \figref{fig:ctrl_fig} represents a true \ac{NEWG} distribution
\figref{fig:ctrl_true} and estimated with the selected query points
\figref{fig:ctrl_estim}. As described in the figure, the optimal parameters are
well-estimated as the ground truth values.

\section{Synthetic Image Generation}
\label{sec:synth}

Next, we evaluate two different camera attributes and their subsequent increase
in burden on capturing an image. In our previous approach~\cite{jkim-2018-icra},
the function evaluation in Bayesian optimization corresponds to an actual frame
acquisition. Despite the reduced function evaluation in using Bayesian
optimization, image acquisition may slow down when a longer exposure time is
applied. To mitigate the cost in the image grab, we propose using synthesized
images instead of capturing images by assigning a target gain value and exposure
time.  Image synthesizing was introduced by Shim~\cite{shim2014auto} who
generated a synthetic image using gamma correction, which transforms image
intensity using a nonlinear transfer function. This synthesis, which is based on
a gamma function, is oriented to generate a natural scene for human vision and
may not be fully realistic.

In this section, we introduce image synthesizing for a target exposure time and
gain value. Specifically, we determine two scale factors $K_t$ (for exposure
time) and $K_g$ (for gain) to be multiplied to the seed image when generating a
synthetic image. Note that our objective was not to generate realistic image but
to evaluate the \ac{NEWG} score of the synthetic image. To leverage the
computational speed, we synthesized the image from a down-sampled seed image by
multiplying the scale factor $K_{synth}$, which is a combination of $K_t$ and
$K_g$. For the scaling factor associated with exposure time, we use \ac{CRF} and
assume a constant irradiance at image acquisition. The scaling factor regarding
gain is rather straightforward in that we can derive the factor directly from
gain.


\subsection{Image Synthesizing using \ac{CRF}}

We start with revisiting \ac{CRF} from machine vision research. The \ac{CRF}
represents the relationship between the sensor irradiance of the camera and the
measured intensity. In the field of computer vision, research has been conducted
on estimating the \ac{CRF} to achieve a high dynamic range image. Assuming that
the CRF is spatially uniform within an image, the \ac{CRF} can be widely
estimated from the same set of images with different but known exposures
\cite{lin2005determining}, \cite{debevec1997recovering}, and \cite{ng2007using}.
Among the existing methods, we focus on \ac{CRF} estimation by
\cite{debevec1997recovering}, who expressed the nonlinear relationship of the
\ac{CRF} as
\begin{equation}
  \label{eq:crf_px}
  I_{x} = f(E_x \Delta t)
\end{equation}
, where $I_x$ is the measured intensity level at the pixel location $x$, $E_x$
is the irradiance of the image at a pixel location $x$, $\Delta t$ is exposure
time.

Note that our objective is to quickly generate synthetic images for an optimal camera
attribute search, not to generate the most similar synthetic images to real. In
that sense, we further simplified \eqref{eq:crf_px} by averaging over the entire
image as
\begin{equation}
  \label{eq:crf}
  I = f(E \Delta t).
\end{equation}
$E$ is now the average irradiance of the image. Although this \ac{CRF} is
unknown, we do know it is monotonic and continuous. Using this function
property, many \ac{HDR} imaging research studies \cite{grossberg2003determining}
focus on the inverse of \ac{CRF}  \eqref{eq:crf}, called an inverse response
function for fitting and image synthesizing. The inverse \ac{CRF} is in a log
form and is a function of irradiance and exposure time as equation below.
\begin{equation}
   g(I) = \ln (E \Delta t).
   \label{eq:icrf}
\end{equation}
%

\begin{figure}[!t]
  \centering%
  \def\width{0.46\columnwidth}%
  \subfigure[Inverse CRF fitting]{
    \includegraphics[trim = 25 260 120 230, clip, width=\width] {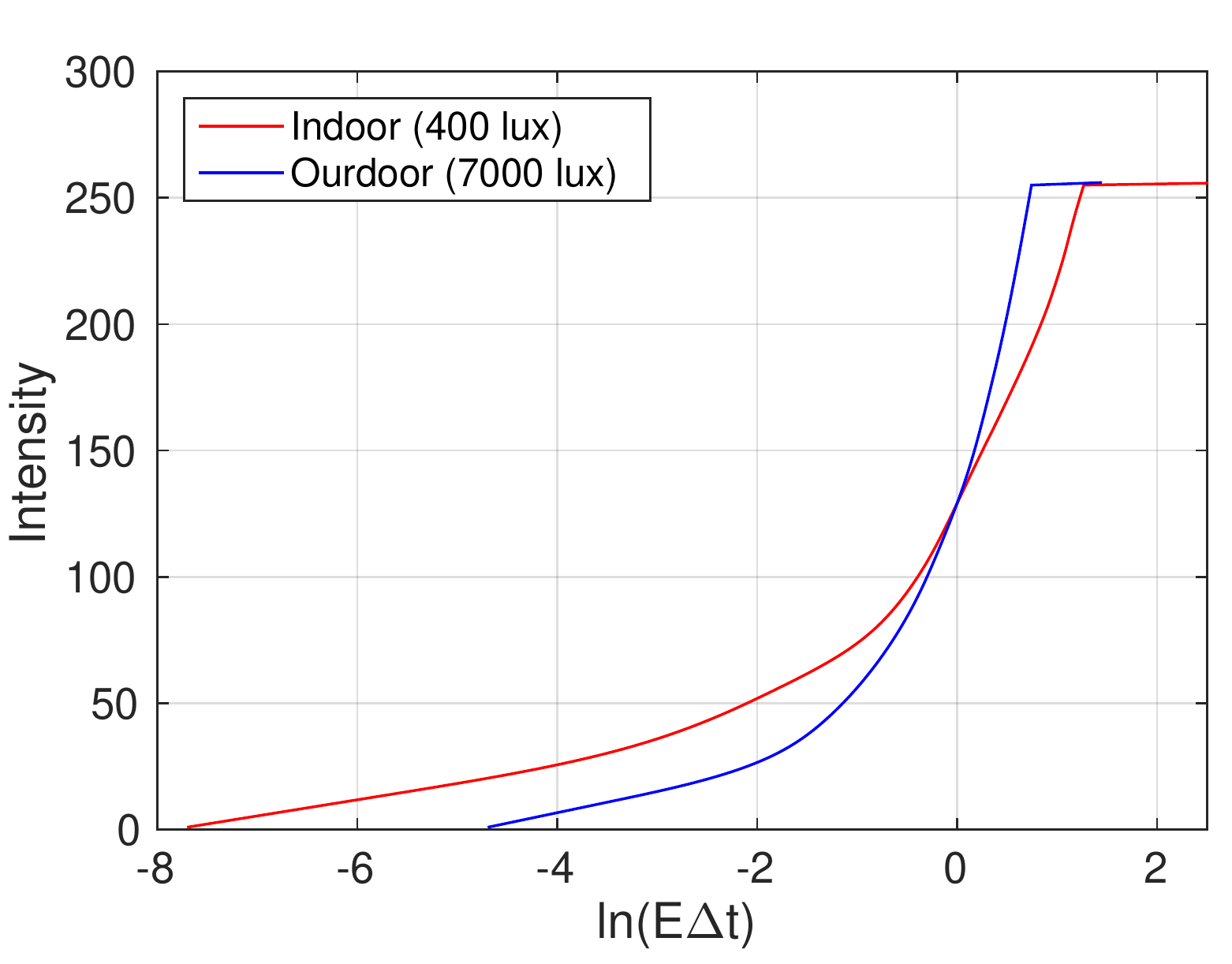} 
    \label{fig:crf_t}
  }
  \subfigure[Synthesizing scale]{
    \includegraphics[trim = 25 260 120 230, clip, width=\width] {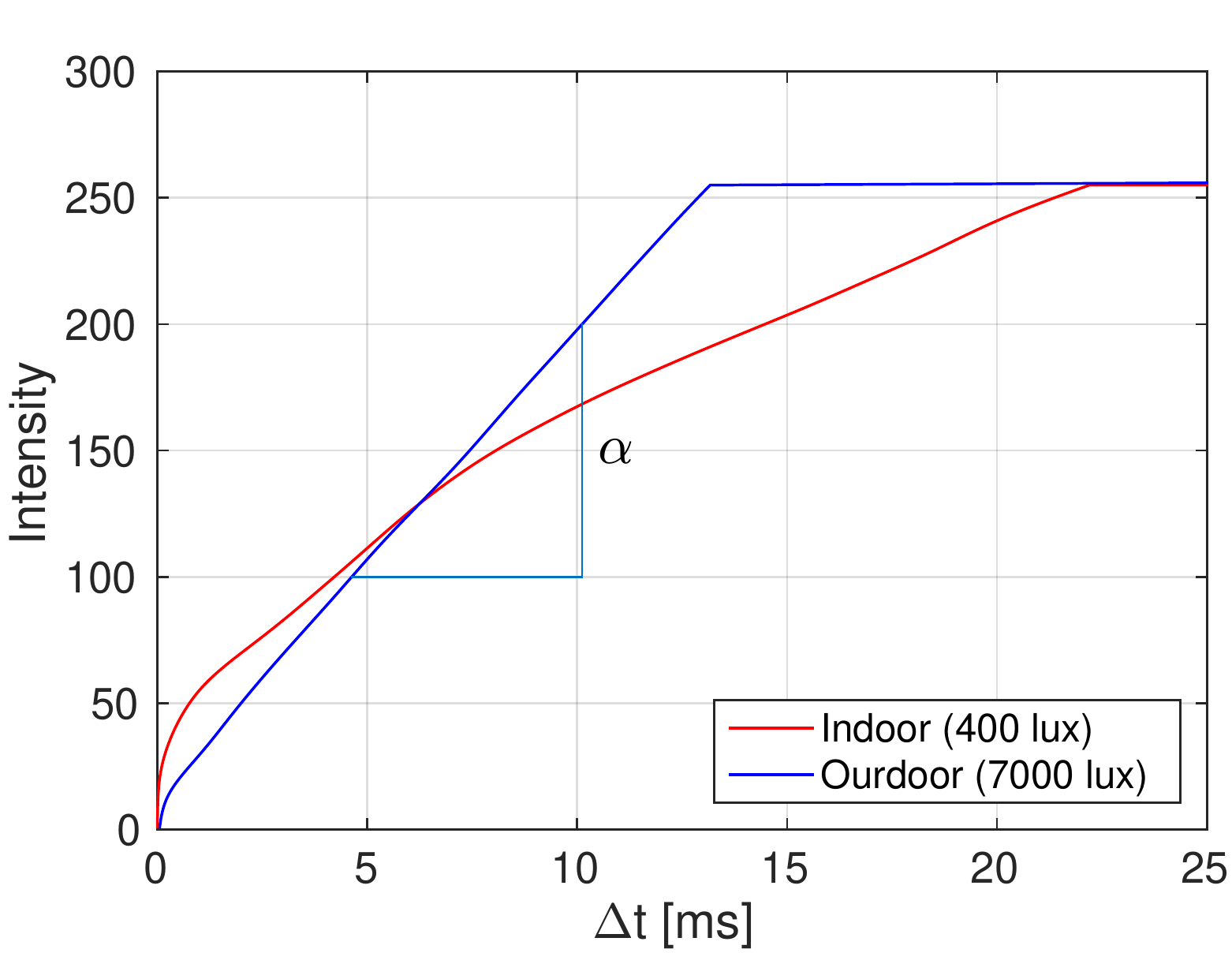} 
    \label{fig:crf_ln_t}
  }
  \caption{The example of inverse camera \ac{RF} for exposure time. The intensity is evaluated
  over randomly sampled pixel points.}
  \label{fig:crf}
  \vspace{-2mm}
\end{figure}

Following \cite{debevec1997recovering}, two image acquisitions fully determines
this inverse \ac{CRF}. Furthermore, these acquisitions are readily available via
an initial function evaluation for Bayesian optimization. To exploit the
function, we need to determine the overall irradiance applied to an image $(E)$
for exposure time respectively by fitting the data to \equref{eq:icrf}. We
follow the same procedure introduced in \cite{debevec1997recovering}. Unlike
conventional methods that evaluate \ac{RF} over a series of images to exactly
fit the function in \equref{eq:crf_px}, our focus is to find $E$ and fit the
\ac{RF} quickly, so as to be used in image synthesis corresponding to arbitrary
camera attributes. For this purpose, we fit the inverse \ac{CRF} using two
sample points at the boundary. For example, minimum and maximum exposure time
are evaluated to generate two samples for the fitting. A sample fitted graph on
exposure time is shown in \figref{fig:crf}.

\figref{fig:crf} presents a fitted inverse \ac{CRF} for two different irradiance
values (i.e., indoor and outdoor). The shape changes as the irradiance changes.
Plotting intensity $g(I)$ with respect to $E\Delta t$ and removing log reveals a
locally linear characteristic of the function as shown in \figref{fig:crf_ln_t}.
Our scaling factor is motivated by this plot and
\cite{grossberg2003determining}. In \cite{grossberg2003determining}, authors
computed a synthetic image at a target exposure time $(\Delta t_2)$ using an
obtained image at the current exposure time $(\Delta t_1)$, inverse \ac{CRF} and
the ratio between $\Delta t_1$ and $\Delta t_2$. When we know the ratio $\gamma
= \Delta t_2 / \Delta t_1$, image intensity corresponds to  $\Delta t_2$ becomes
\begin{equation}
  I_{\Delta t_2} = g^{-1}\left( \gamma g (I_{\Delta t_1}) \right) = g^{-1}(\gamma g(\ln E + \ln \Delta t_1)).
\end{equation} 
Using this intensity relation, for target exposure time $\Delta t_{s}$, we determine scale
factor $K_t$ to be applied for each pixel of the seed image ($I_o$) captured
with the seed exposure time $(\Delta t_o)$. Using at least two real images obtained, we fit
the inverse \ac{CRF} and compute the intensity ratio from the plot as in
\figref{fig:crf_ln_t}. Assuming the irradiance ($E$) of the image is constant,
the intensity ratio can be used to predict the average intensity of the
synthesized image $I_s$ as
\begin{eqnarray}
  \label{eq:alpha}
  \alpha &=& \frac{g(I_s)- g(I_o)}{\Delta t_s - \Delta t_o}\\
  K_t = \frac{g(I_s)}{g(I_o)} &=& \frac{\alpha (\Delta t_s - \Delta t_o)+g(I_o)}{g(I_o)}
\end{eqnarray}
, where $g(I_s)$ is the average intensity of the synthesized image found in the inverse \ac{CRF} function. Once fitted with an inverse \ac{CRF}, the scaling factor
$K_t$, which is associated with exposure time is directly obtained.

\begin{figure}[!t]
 \centering%
  \def\width{0.45\columnwidth}%
  \subfigure[Indoor (\unit{400}{lux})]{
     \includegraphics[trim = 80 230 120 230, clip, width=\width] {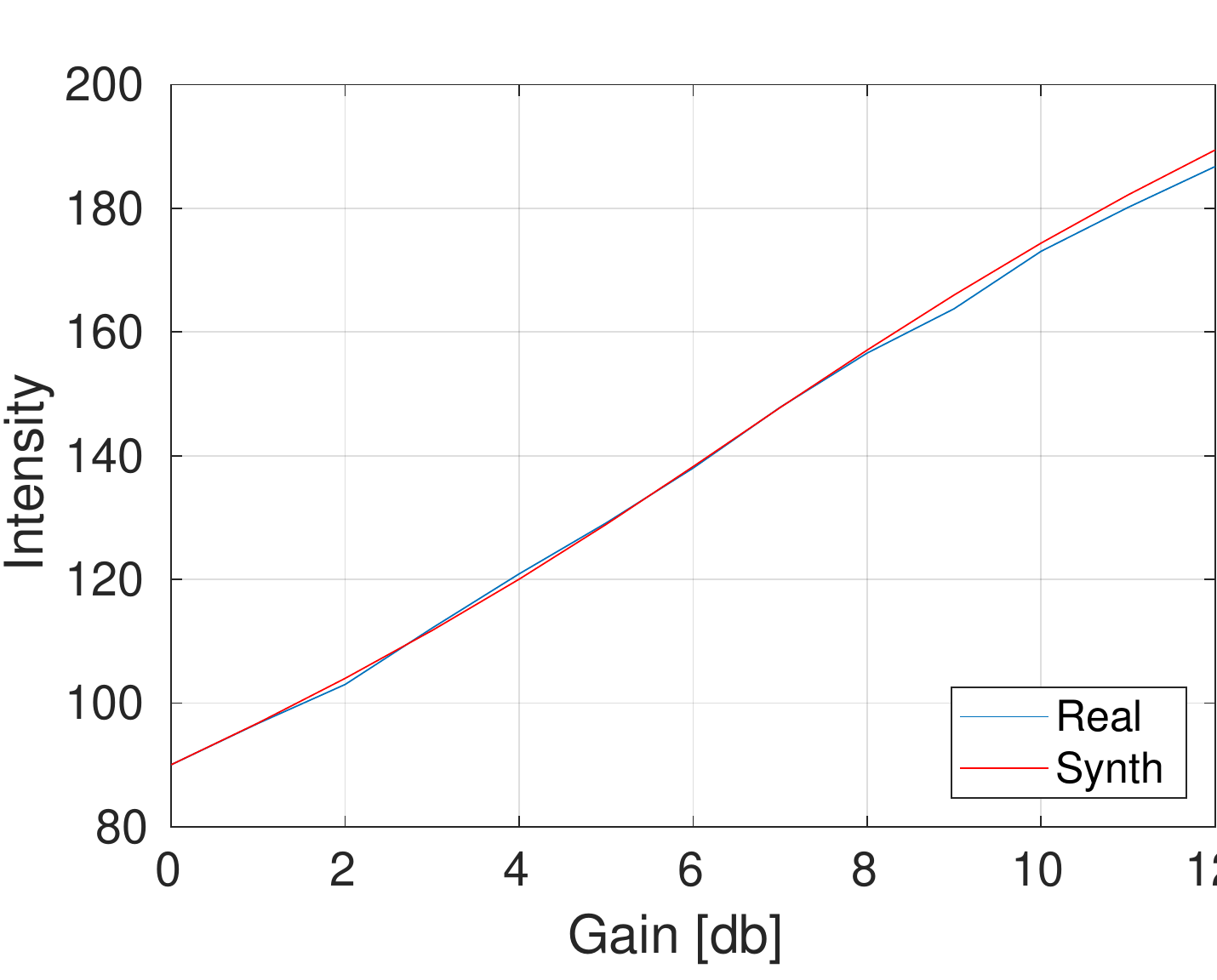} 
     \label{fig:fit_gain_in}
  }\hfill
  \subfigure[Outdoor (\unit{7000}{lux})]{
     \includegraphics[trim = 80 230 120 230, clip, width=\width] {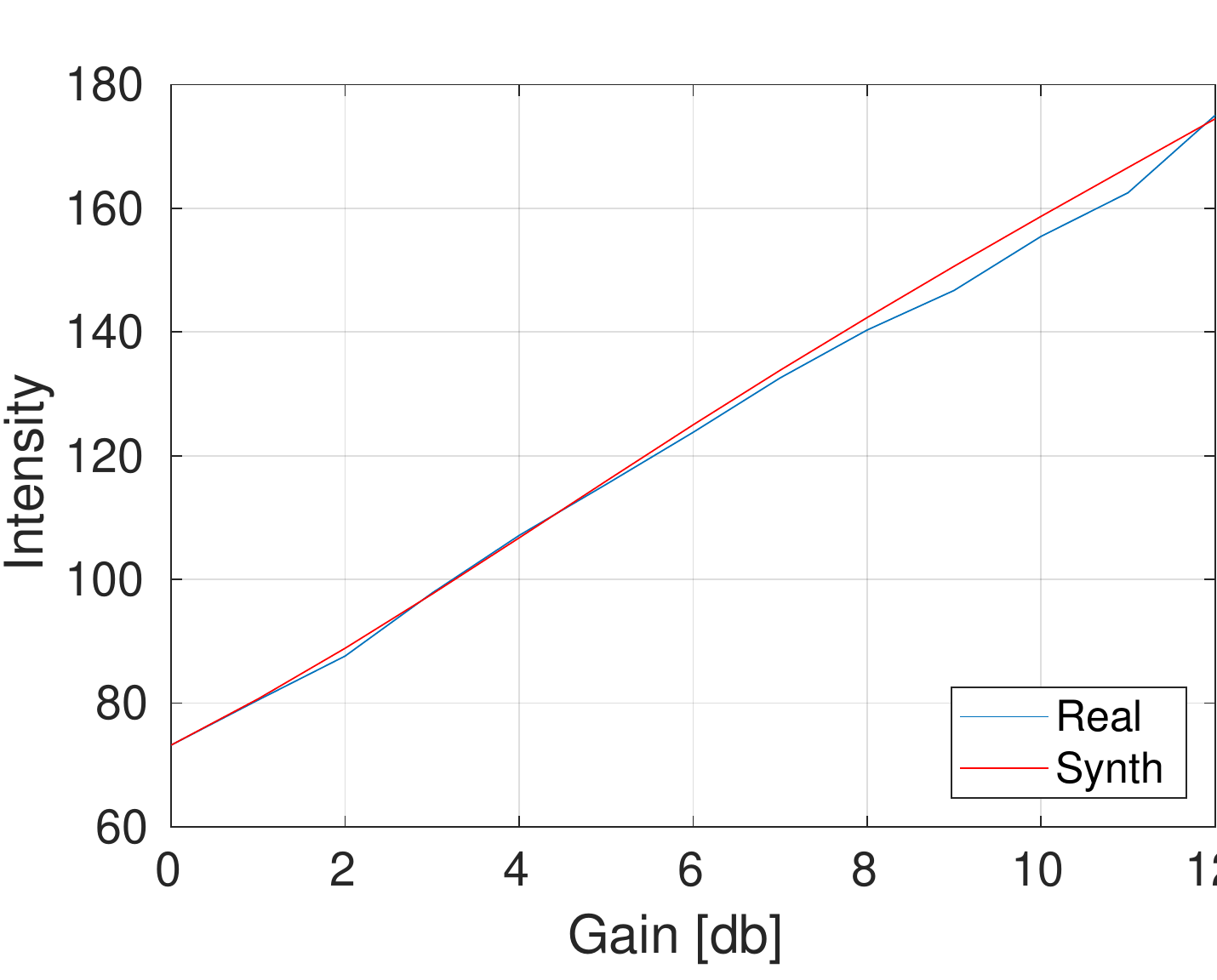} 
    \label{fig:fit_gain_out}
  }

  \caption{The average image intensity of the real and synthetic image are
  plotted while changing the gain. Images are synthesized via \eqref{eq:kg} by
  using the image intensity captured in \unit{0}{dB} as the seed image ($I_o$).}

  \label{fig:gain_fit}
  \vspace{-2mm}
\end{figure}

\begin{figure*}[!t]
  \scriptsize
  \centering

  \begin{minipage}{0.7\textwidth}
  \subfigure[Comparison of optimal parameters]{
    \label{table:metric_eval}
    \begin{tabular}{cc|ccccccc}
      \multicolumn{2}{c|}{Illuminance [lux]}              & 10                  & 50                   & 80               & 100               & 320             & 400                & 1000\\
      \hline
      \multicolumn{2}{c|}{Environment}                    & \textit{Dark scene} & \textit{Living room} & \textit{Hallway} & \textit{Overcast} & \textit{Office} & \textit{Sunny} & \textit{Artificial light}\\
      \hline
      SNR-ignored                    & $\Delta t^*$ [ms]  & 17.0                 & 12.0             & 5.5              & 5.5               &   2.5           & 2.0                & 2.0\\
      (EWG)                          & $G^*$ [dB]         & 11                   & 3                & 0                & 0                 &  11             & 4                  & 1\\
      \hline
      SNR-considered                 & $\Delta t^*$ [ms]  & 19.5                 & 12.5             & 5.5              & 5.0               &   7.0           & 2.0                & 1.5\\
      (NEWG)                         & $G^*$ [dB]         & 9                    & 0                & 0                & 0                 &  1              & 4                  & 1\\
      \hline \vspace{1mm}
    \end{tabular}\\
  }
  \end{minipage}
  \hfill
  \begin{minipage}{0.25\textwidth}
    \centering
    \subfigure[Camera rig]{
    \label{fig:system}
    \includegraphics[width=0.7\columnwidth] {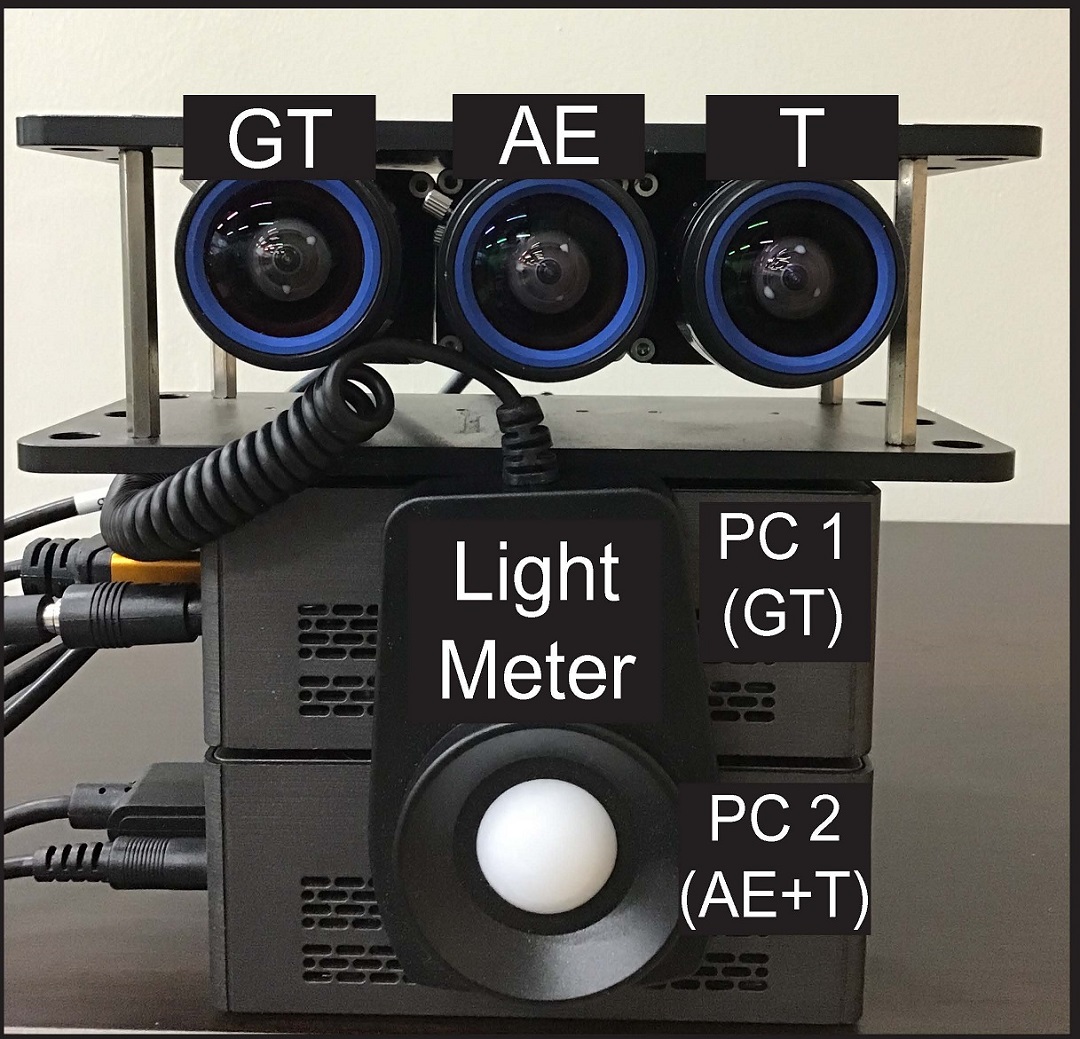}
    }
  \end{minipage}

  \caption{\subref{table:metric_eval} comparison of optimal parameters ($\Delta
  t^*$ and $G^*$) depending on the choice of metric (i.e., SNR-ignored (EWG) vs.
  SNR-considered (NEWG)). \subref{fig:system} A multi camera rig and a light
  meter used in the test.  Each camera operates controller using gain and
  $\Delta t$ (\texttt{GT}), auto exposure (\texttt{AE}), and exposure time only
  (\texttt{T}) respectively.}

\end{figure*}


\subsection{Image Synthesizing for Gain}

Determining scale factor $K_g$ for gain is straightforward using the definition
for the gain, $g = 20 \log_{10} \frac{g(I_s)}{g(I_o)}$. Because we wish to
synthesize an image with gain factor $K_g$, the above equation can be further
simplified as
\begin{equation}
  \label{eq:kg}
  g(I_s) = K_g g(I_o) = (10 / F_n)^{\frac{g}{20}}g(I_o) \simeq 7.01^{\frac{g}{20}}g(I_o)
\end{equation}

Factoring with $F_n = \sqrt{2}$ was empirically determined. For indoor and outdoor
evaluation dataset, we plotted the average intensity of both real and synthetic
images by varying gain as in \figref{fig:gain_fit}. As can be seen in the
figure, the scaling factor successfully captures the average image intensity for
both indoor and outdoor settings.

\section{Experiments}
\label{sec:exp}

To evaluate the proposed metric and image synthesis capability, we performed an
exhaustive evaluation of data collection in various light conditions. The
summary of the data is provided in the table in \figref{table:metric_eval}. This
image synthesizing is included into our control scheme to produce a fast
evaluation for optimal camera attribute control.

\subsection{Validation of Image Evaluation Metric}
\label{sec:data}

We first validate the image evaluation metric by comparing our previous metric
against the proposed metric that additionally considers the \ac{SNR}.  When gain
is simultaneously controlled with exposure time, the \ac{SNR} affects image
quality and should be incorporated into the metric.

In various illumination conditions, we collected an exhaustive dataset to
validate our metric and image synthesis quality. We captured actual images while
changing exposure time from \unit{1}{ms} to \unit{20}{ms}. For this test, this
range was selected because the image blurs when the exposure time is too large.
Similarly, actual images with various gain values were obtained by changing the
gain values from \unit{0}{dB} to \unit{12}{dB}. For each pair of exposure time
and gain, an image was captured and an associated metric was evaluated.

\begin{figure}[!t]
  \centering%
  \def\width{0.29\columnwidth}%
  \centering%
  \subfigure[10 lux]{
    \includegraphics[trim = 90 240 90 280, clip, width=\width]{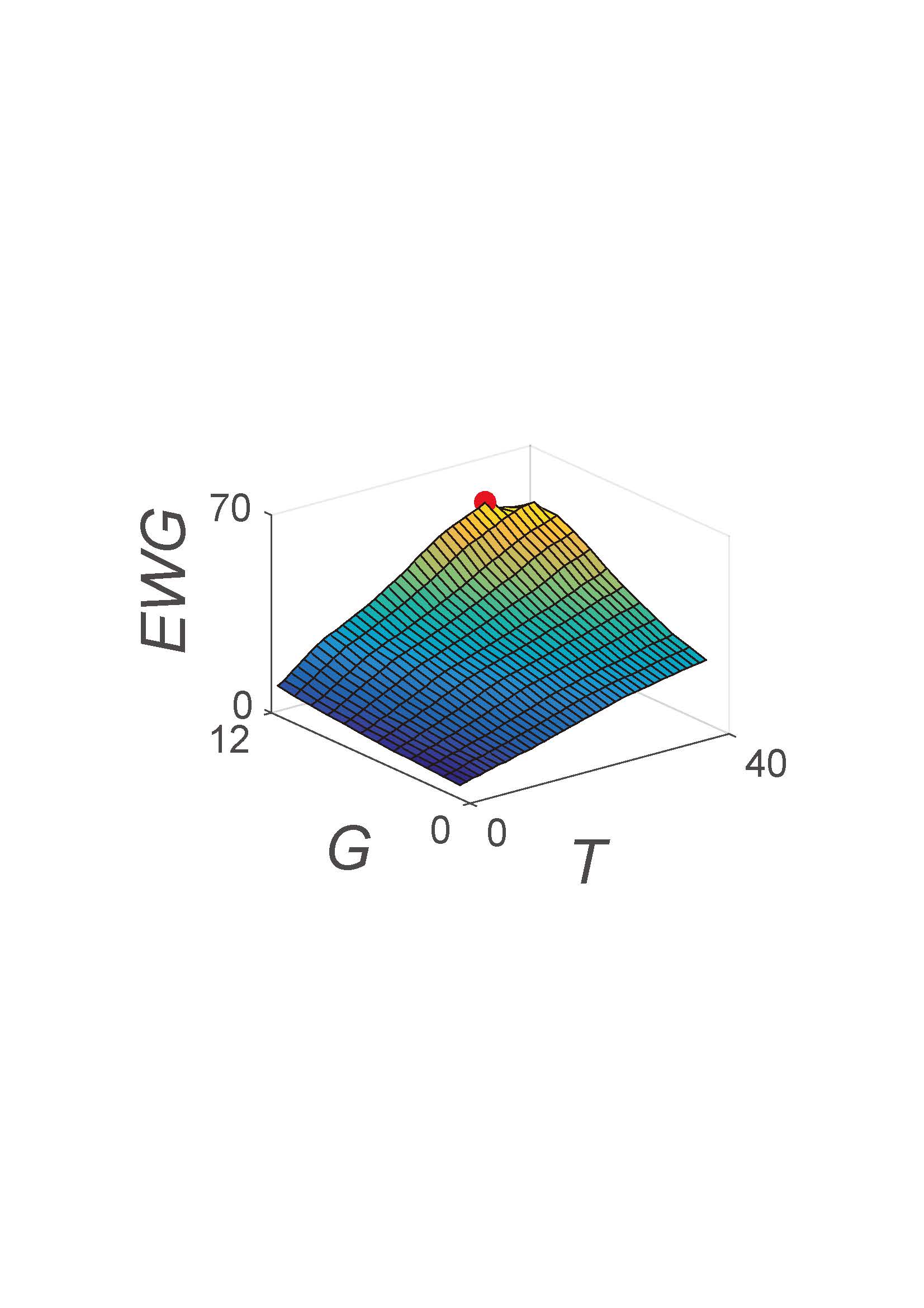}
    \label{fig:env1}
  }
  \subfigure[320 lux]{
    \includegraphics[trim = 90 240 90 280, clip, width= \width] {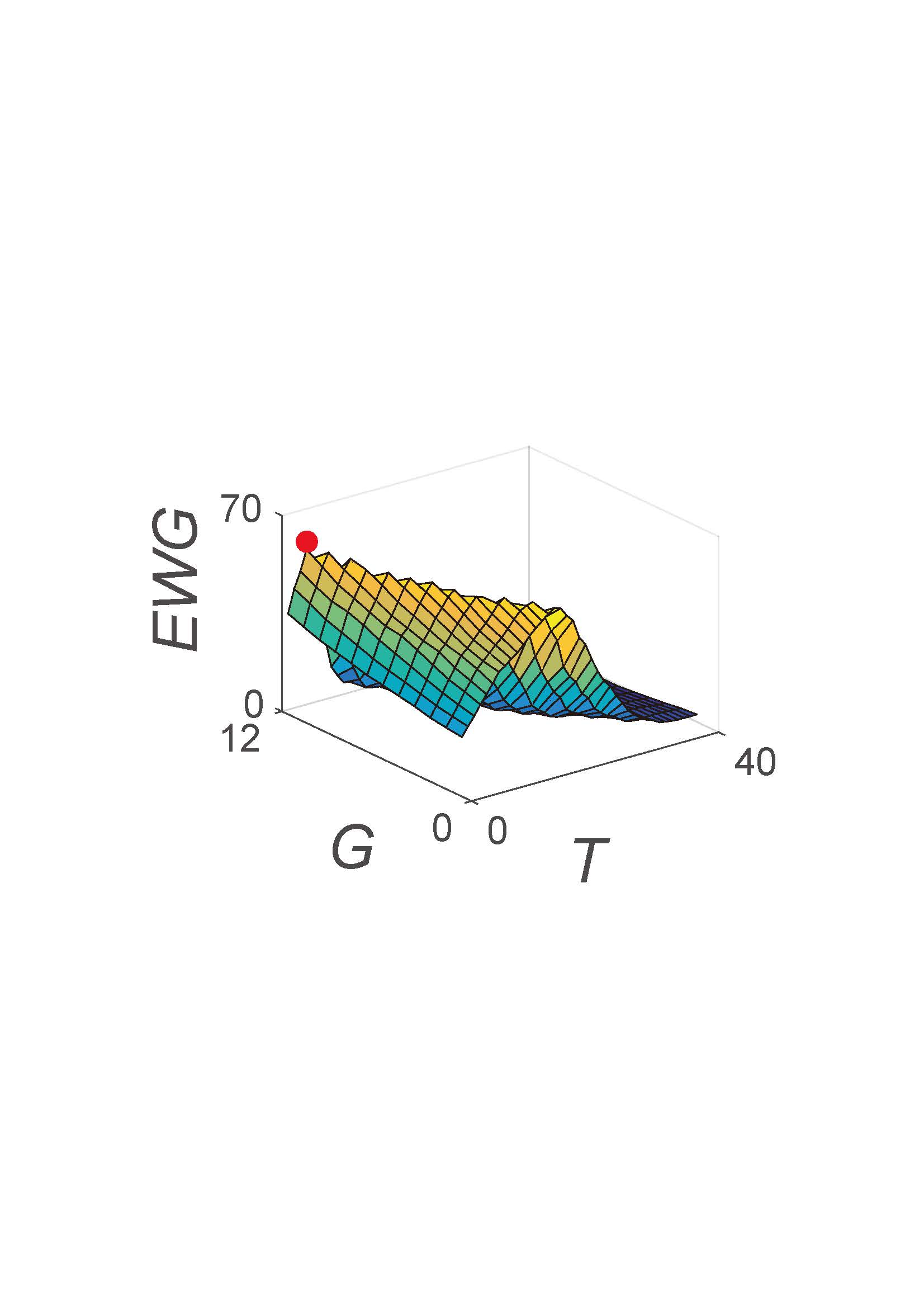}
    \label{fig:env5}
  }
  \subfigure[1000 lux]{
    \includegraphics[trim = 90 240 90 280, clip, width=\width] {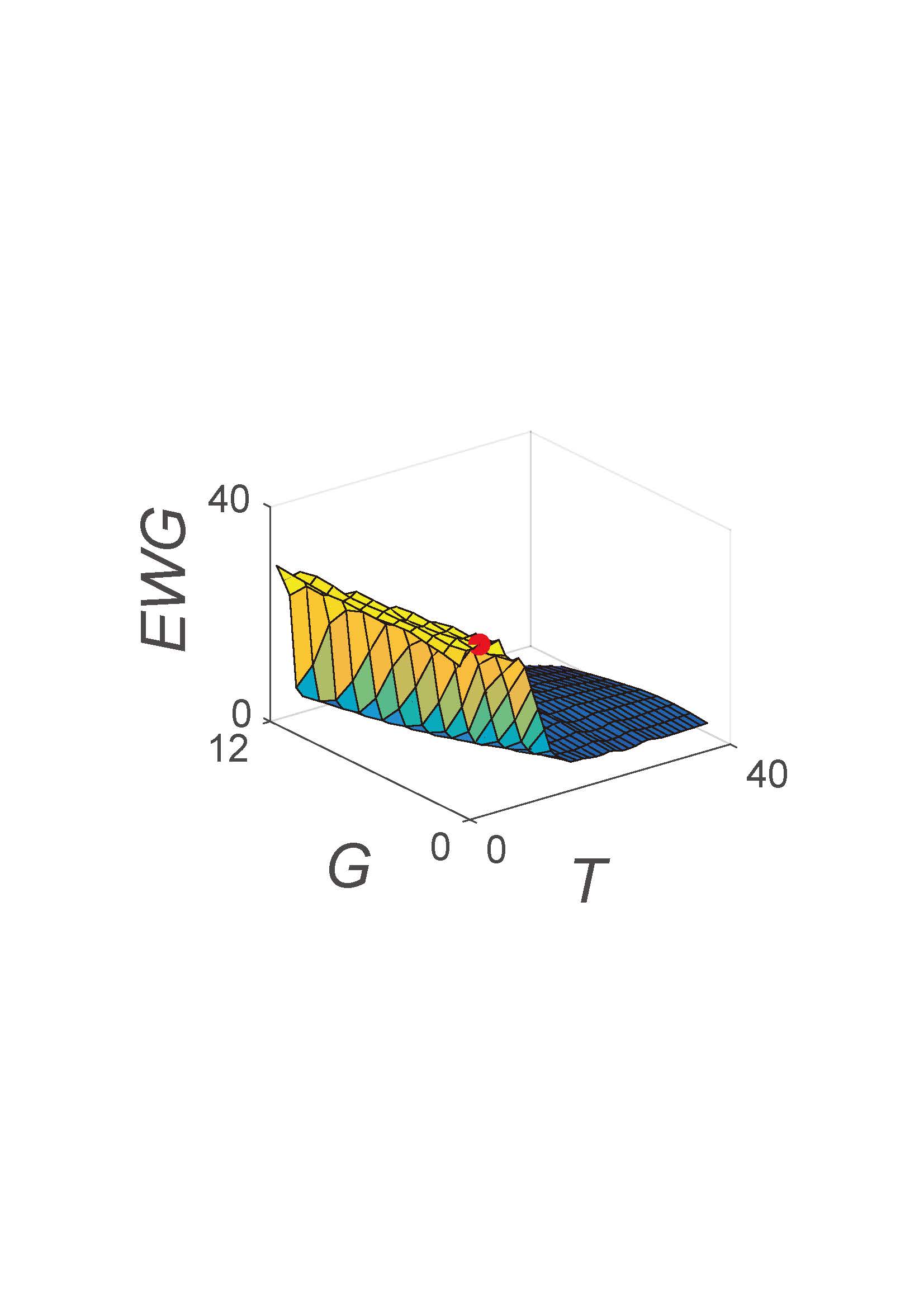}
  \label{fig:env7}
  }\\
  \subfigure[10 lux]{
    \includegraphics[trim = 90 240 90 280, clip, width=\width]{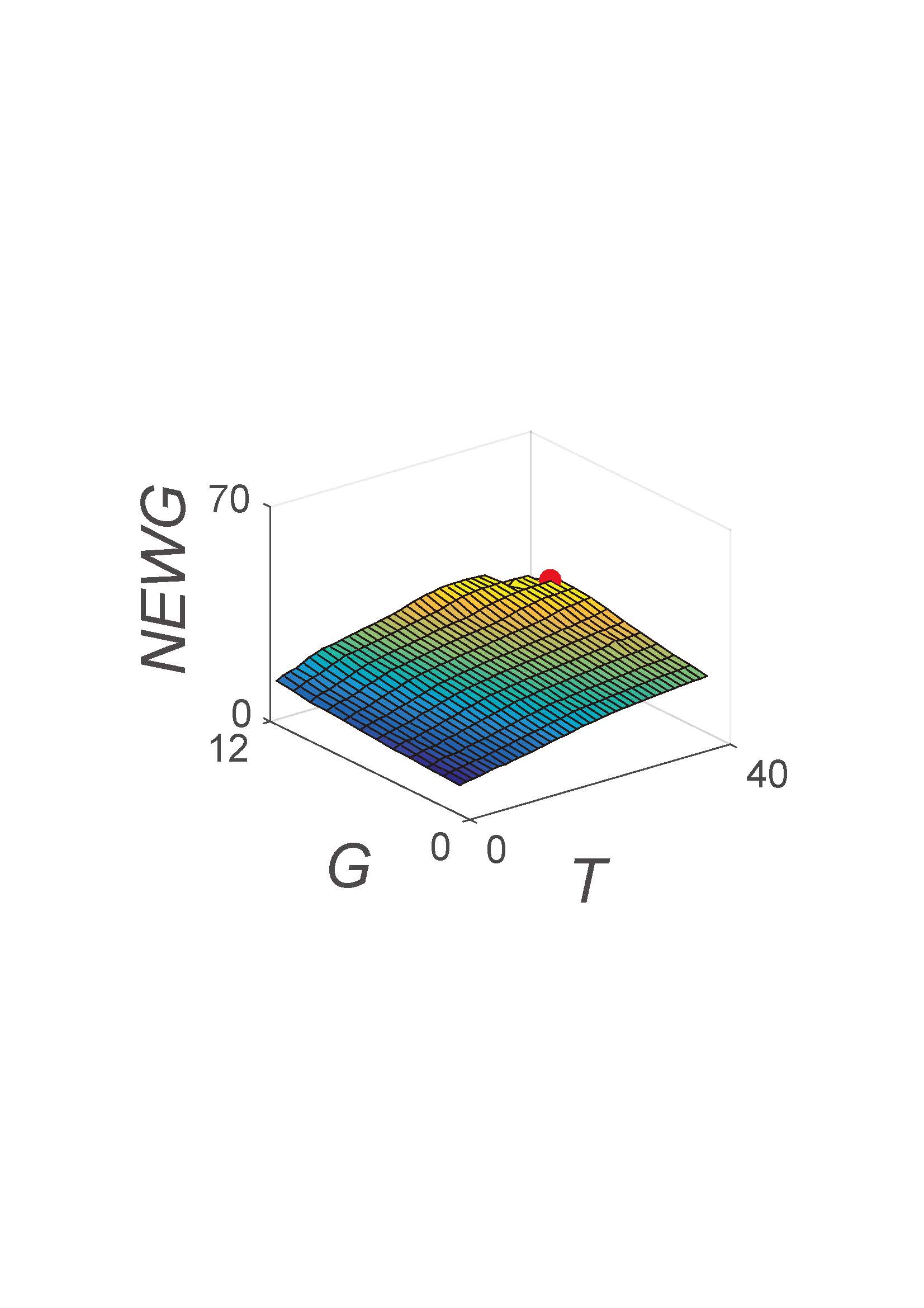}
    \label{fig:gain1}
  }
  \subfigure[320 lux]{
    \includegraphics[trim = 90 240 90 280, clip, width=\width] {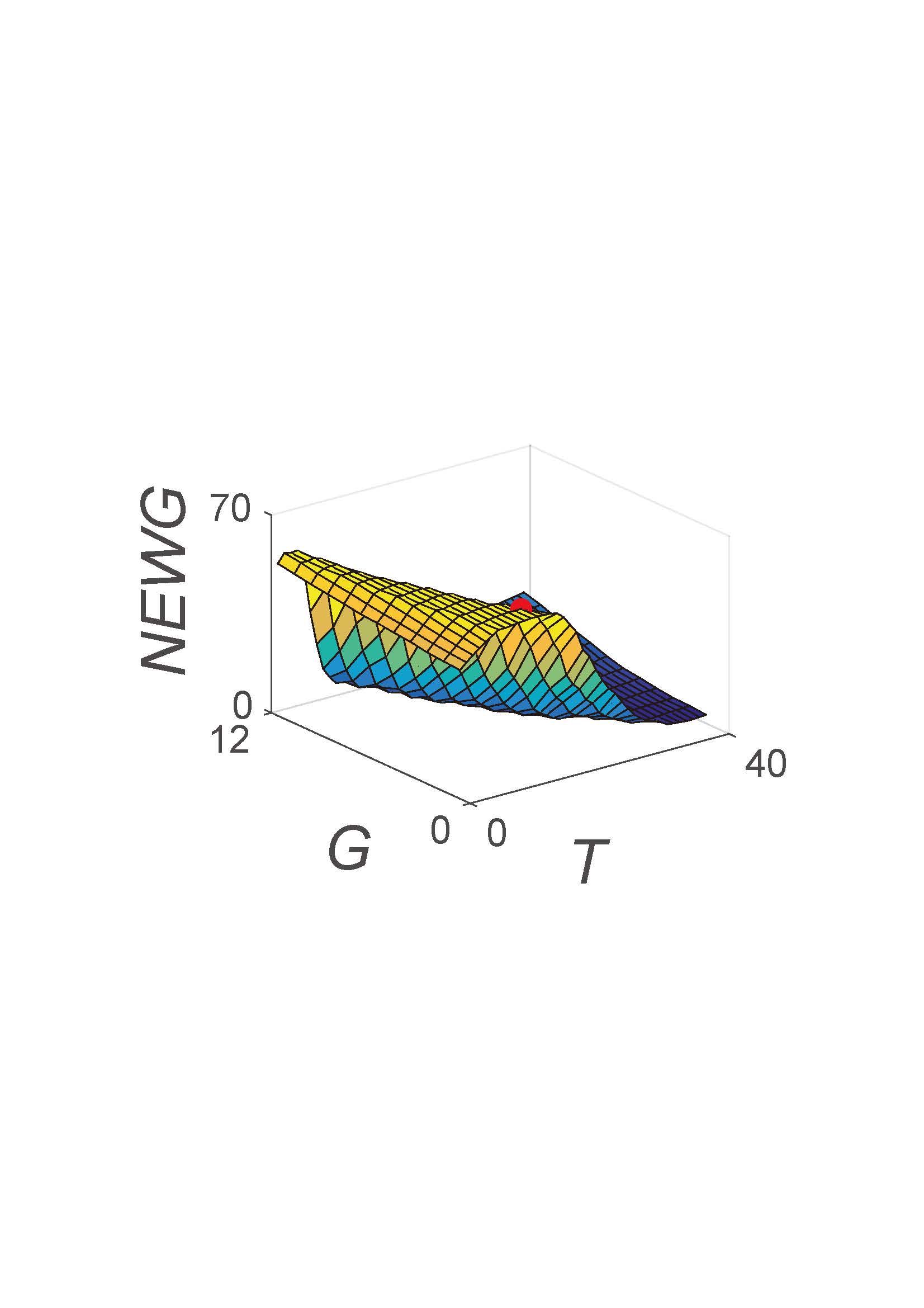}
    \label{fig:gain5}
  }
  \subfigure[1000 lux]{
    \includegraphics[trim = 90 240 90 280, clip, width=\width] {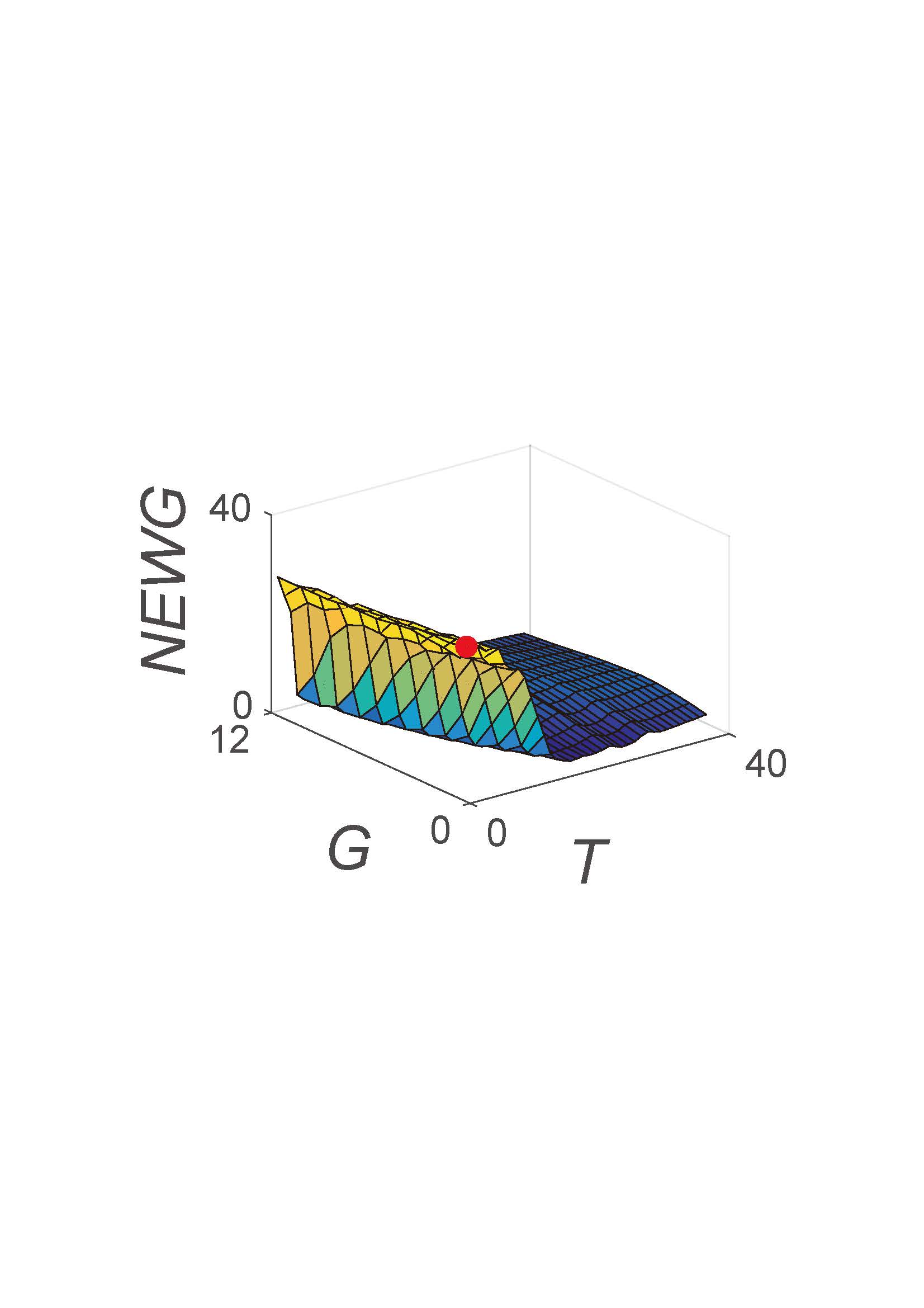}
    \label{fig:gain7}
  }

\caption{Sample image evaluation metric plotted with respect to the gain and
exposure time. Top row depicts the metric when only considering saturation and
the bottom row represents the metric when both saturation and the \ac{SNR}. The
red dot represents the optimal value of each metric. Note the change in the
optimal values for 320 lux when SNR is additionally considered.}

  \label{fig:metric_eval}
\end{figure}

\figref{fig:metric_eval} compares two different metrics while changing exposure
time and gain exhaustively. The newly proposed \ac{SNR}-considering metric
(bottom row) was plotted in comparison to a metric that does not consider the
\ac{SNR} (top row). Unlike the metric without the \ac{SNR} in our earlier work
\cite{jkim-2018-icra}, the new metric effectively compensates for noise even in
dark environments. As can be seen in the dark case (10 lux), the overall curve
of the graph increases less drastically with the new metric (\figref{fig:gain1}
and \figref{fig:env1}). By including noise into the metric, higher exposure and
gain are less favored by the penalty from the increase in noise. The previous
metric heavily aimed to eliminate saturation areas in the images (e.g., windows)
obtained from an office environment (320 lux). However, the new metric was
willing to tolerate saturation provided that no higher noise is yielded in the
images (\figref{fig:gain5}). In addition, this metric prevents the image from
creating an excessively high gain, even in a bright environment that can be
sufficiently controlled by exposure time only. On the other hand, there is
little effect in very bright environments (\unit{1000}{lux}). This is because
our algorithm searches for the balanced exposure time and gain. In a dark
environment, the image quality is hardly controlled only by the exposure time
control.

Table in \figref{table:metric_eval} lists the optimal exposure time and optimal
gain for each environment determined by two metrics. As can be seen in the
table, the \ac{SNR}-considered metric penalized the gain values effectively,
preventing an optimal high gain value to be reached. For example, when the
environment is fairly bright (\unit{320}{lux}), images with a low exposure time
(\unit{2.5}{ms}) but high gain (\unit{11}{dB}) were preferred because they
contained more gradients despite the increased noise. On the other hand, when
the \ac{SNR} is considered, this issue is alleviated by selecting a larger
exposure time (\unit{7.0}{ms}) and low gain (\unit{2}{dB}). As similar
improvement was found when the environment was dark (\unit{10}{lux}) forcing the
optimal value to occur with a lower gain value.



\begin{figure*}[tb]
  \centering%
  \subfigure[Indoor evaluation. The figures show images obtained at different
  exposure times in \unit{1}{ms} intervals during the day (left top), images
  measured differently in \unit{1}{dB} intervals (right top), and synthetic
  images corresponding to those images (bottom).]{
    \includegraphics[width=15.5cm,height=9.3cm, ]{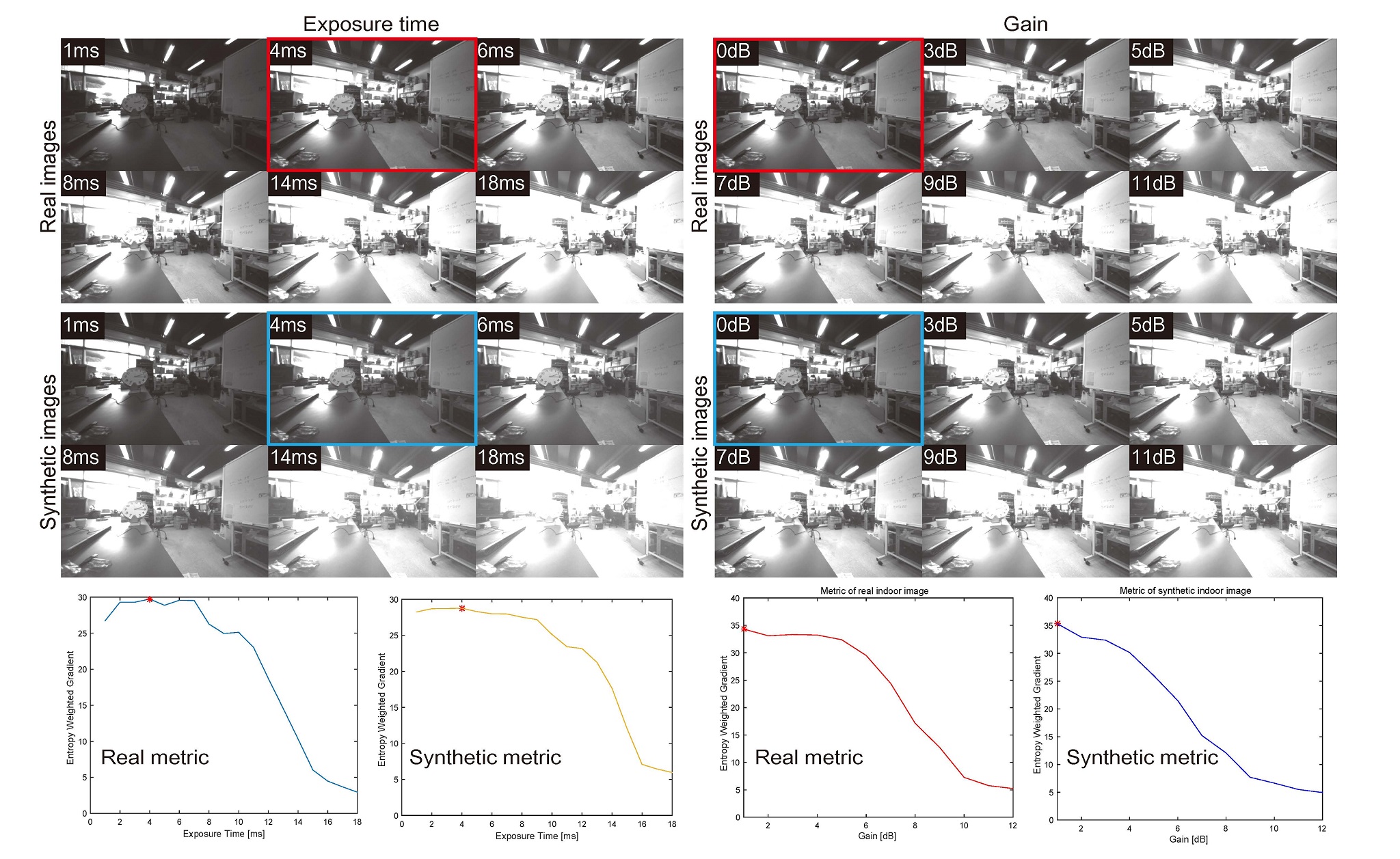}
    \label{fig:indoor_eval} \vspace{-1mm}
  }
  \subfigure[Outdoor evaluation. The figures show images obtained at different
  exposure times in \unit{5}{us} intervals during the day (left top), images
  measured differently in \unit{1}{dB} intervals (right top), and synthetic
  images corresponding to those images (bottom).]{
    \includegraphics[width=15.5cm,height=9.3cm, ]{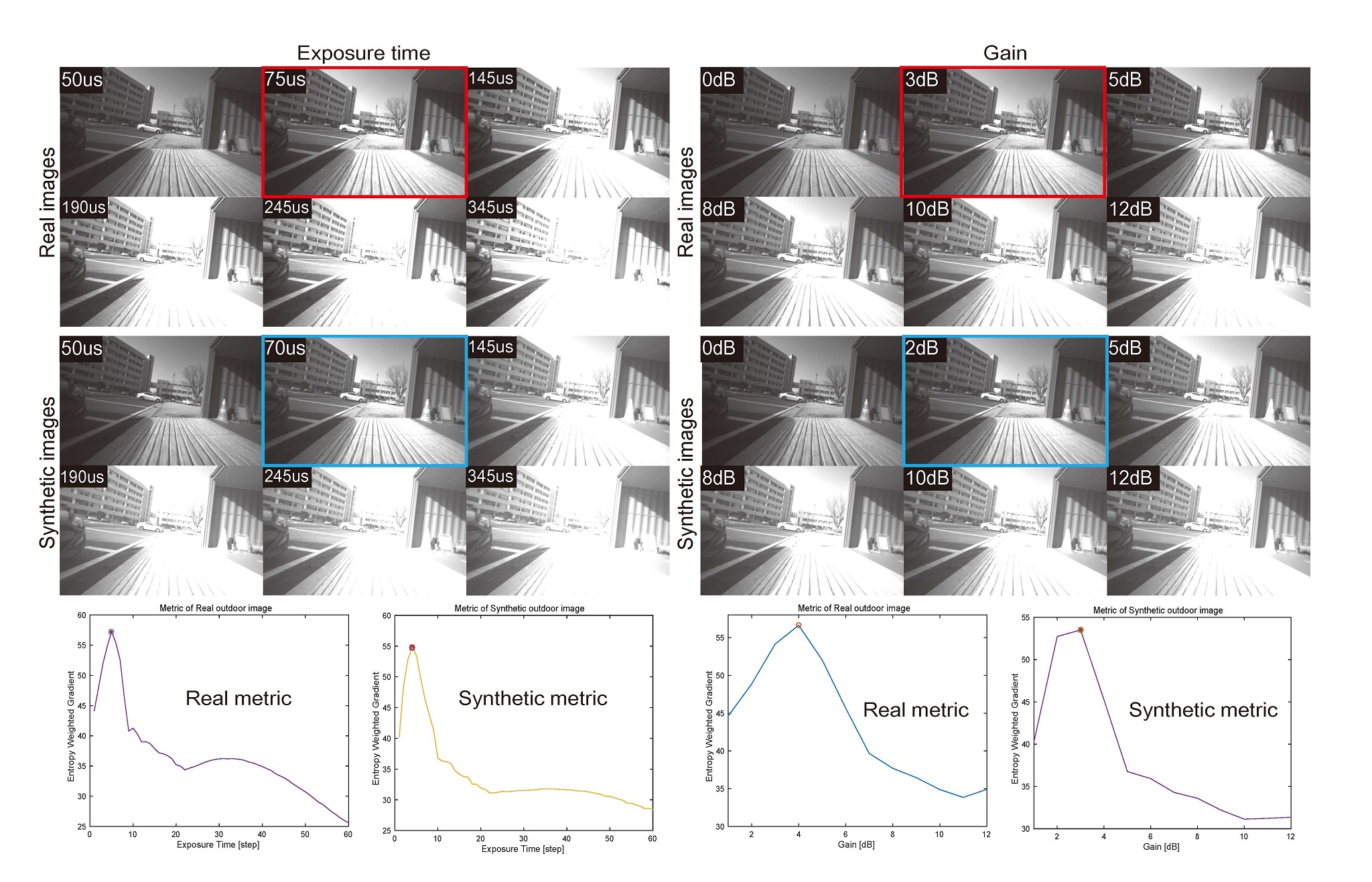}
    \label{fig:outdoor_eval}
  }
    \caption{Synthesized image evaluation. Comparing and analyzing synthetic image for camera gain and synthetic image  for exposure time using camera response function. The red box is the image with the maximum amount of information (NEWG) in the actual image, and the blue box is the image that shows the maximum amount of information in the synthetic image.}
    \vspace{-3mm}
\end{figure*}

\subsection{Validation of Image Synthesis}

\begin{figure*}[!t]
  \centering%

  \subfigure[Indoor test]{%
    \includegraphics[width=0.9\textwidth] {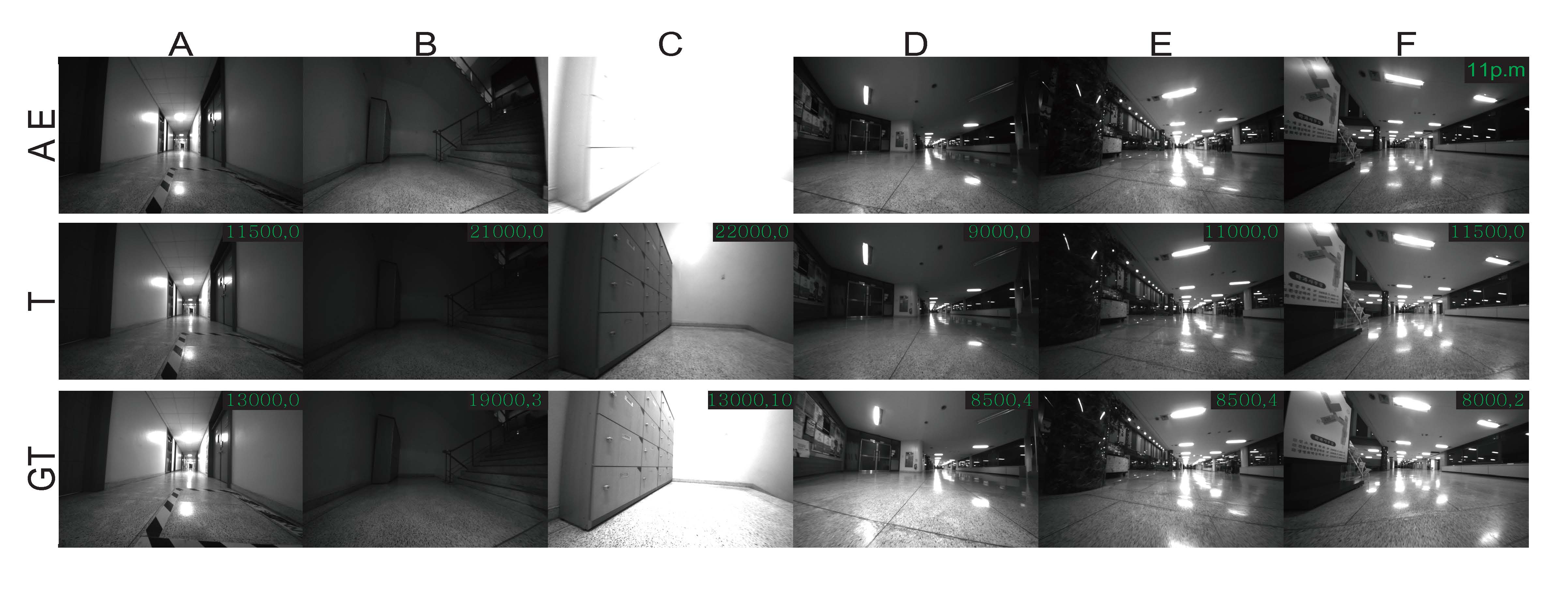}
    \label{fig:ctrl_test_in}
  }\\
  \subfigure[Outdoor test]{
    \includegraphics[width=0.9\textwidth] {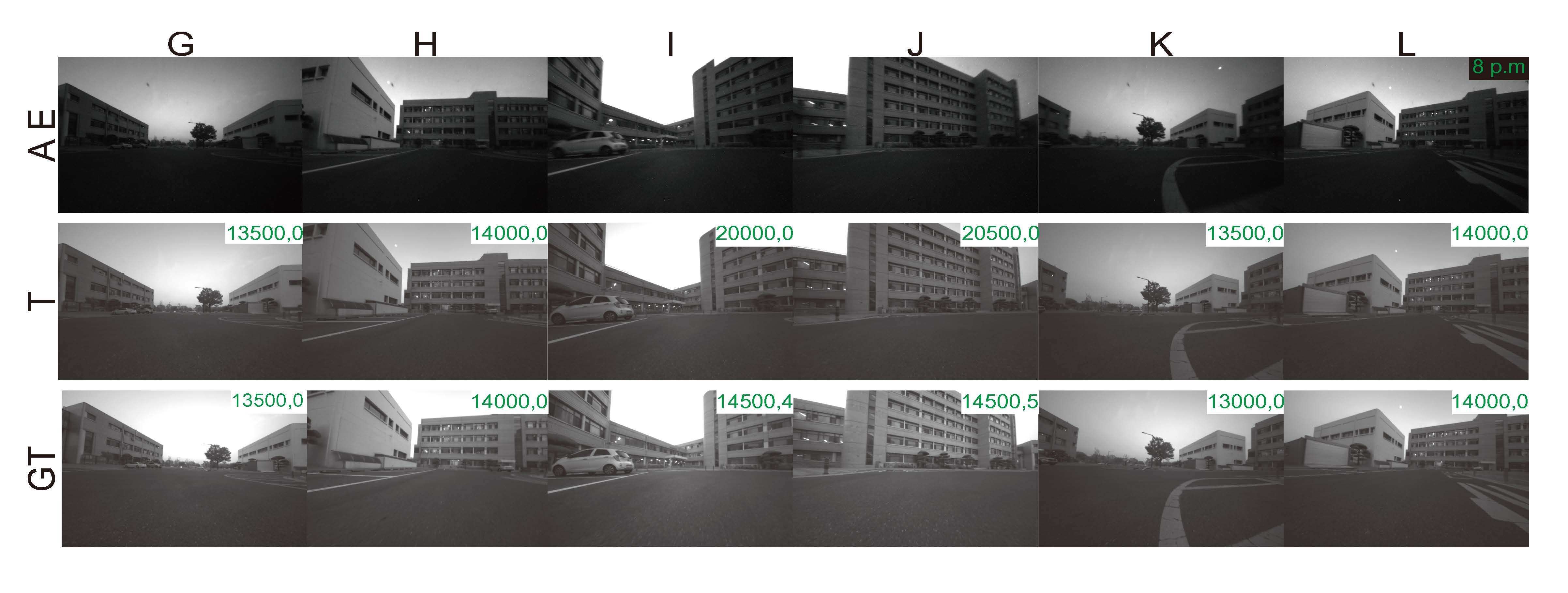}
    \label{fig:ctrl_test_out}
  }

  \caption{For both indoor and outdoor tests, each row represents an acquired
  image that was controlled for automatic exposure (\texttt{AE}), exposure time
  (\texttt{T}), and exposure time and gain (\texttt{GT}).
  \subref{fig:ctrl_test_in} The experiment was conducted in an indoor
  environment at night. Exposure time (in ms) and gain (in dB) are marked in the
  top right corner of each image. During the drastic change between \texttt{`B'}
  and \texttt{`C'}, \texttt{AE} resulted in saturation for a bright scene
  (\texttt{`C'}) by focusing on the exposure for the dark scene in
  (\texttt{`B'}).  Exposure-only control was insufficient to overcome the dark
  scene (\texttt{`B'}).  \subref{fig:ctrl_test_out} The experiment was conducted
  in an outdoor environment after sunset. Large exposure time for a dark scene
  was not sufficient for \texttt{AE} and \texttt{T}. Due to this limited light,
  motion blur and high noise occurred.}

  \label{fig:ctrl_test}
\end{figure*}

Before applying the synthetic image in our control loop, we validated the
effectiveness of the proposed image synthesis by comparing them with the real
images, which was captured while varying exposure and gain respectively. Using
these actual images as a baseline (top group in \figref{fig:indoor_eval} and
\figref{fig:outdoor_eval}), we synthesized the images for the intended exposure
time and gain values (bottom group in \figref{fig:indoor_eval} and
\figref{fig:outdoor_eval}). For the exposure time based synthesis, we used the
seed image with a minimum exposure time (\unit{1}{ms}) to generate other images.
For the gain based synthesis, we used zero gain (\unit{0}{dB}) image as the seed
image.

We conducted more quantitative analysis by plotting the image quality metric
variation. We measured the proposed image metric from both synthetic and real
images. We would like to stress that our intention was not to synthesize an
image to be the same as a real image. Rather, we intend to achieve an image
synthesis process that follows a similar metric evaluation. As shown in the
metric plots in \figref{fig:indoor_eval} and \figref{fig:outdoor_eval}, the
resulting synthesized image sequence revealed the similar image evaluation
metric when compared to the metric obtained from the real image sequence. Most
importantly, the selected optimal value is the same for both the synthetic and
real image sequences, thus proving that the proposed synthetic images can be
used to find the optimal camera attribute.

\subsection{Camera Attribute Control}

Next, we present experimental validation in both indoor and outdoor
environments. For both tests, three cameras are mounted as in
\figref{fig:system}. For indoor validation, the test was performed in an indoor
corridor at night, where the light was automatically controlled by passenger
motion. As the robot moved, the light turned on and off repeatedly. For outdoor
validation, images were captured right after sunset when the light is limited.




\figref{fig:ctrl_test} summarizes the experimental results from the indoor
environment. When the exposure was controlled automatically (\texttt{AE}), too
large exposure time was assigned for a dark scene, and producing an over-exposed
image in the upcoming frame (first row, \texttt{B} and \texttt{C}). If we only
control exposure time as in our previous study, an under-exposed image occurred
in dark environments (second row, \texttt{B}). Similarly for outdoor,
\figref{fig:ctrl_test_out} shows an outdoor environmental experiment right after
sunset. As it became dark, the auto attribute control (\texttt{AE}) yielded
images with high noise by assigning a high gain value. Furthermore, lack of
light and the large exposure resulted in a motion blur. For both experiments,
the proposed camera attribute control simultaneously controls gain and exposure
time to prevent motion blur and saturation in the entire image.

\begin{figure}[!t]
  \centering%
  \subfigure[Trajectory]{
    \includegraphics[width=0.31\columnwidth] {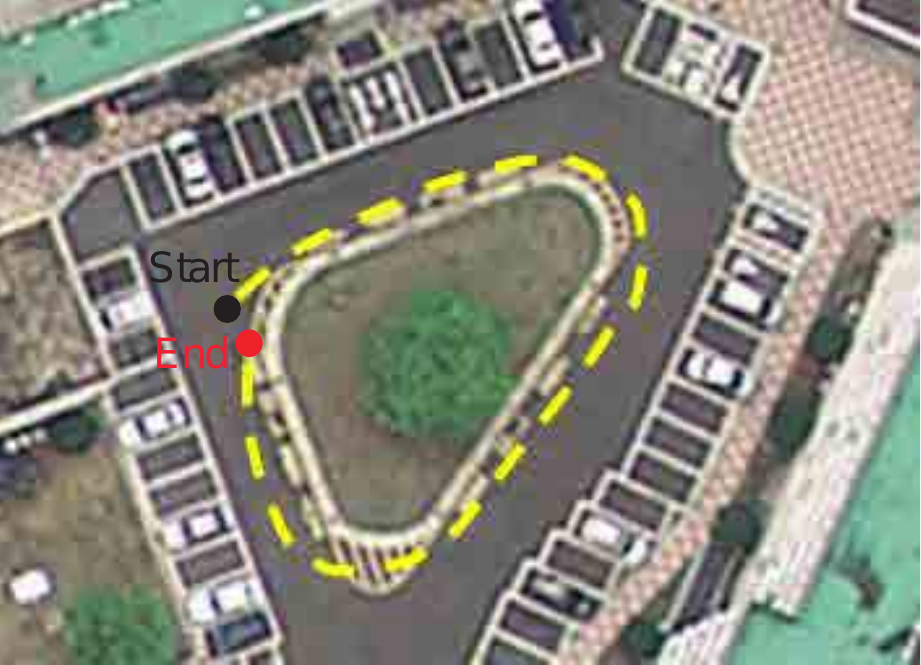} 
    \label{fig:out_airial}
  }%
  \subfigure[$\Delta{t}$ control]{
    \includegraphics[trim = 139 295 122 270, clip, width=0.3\columnwidth] {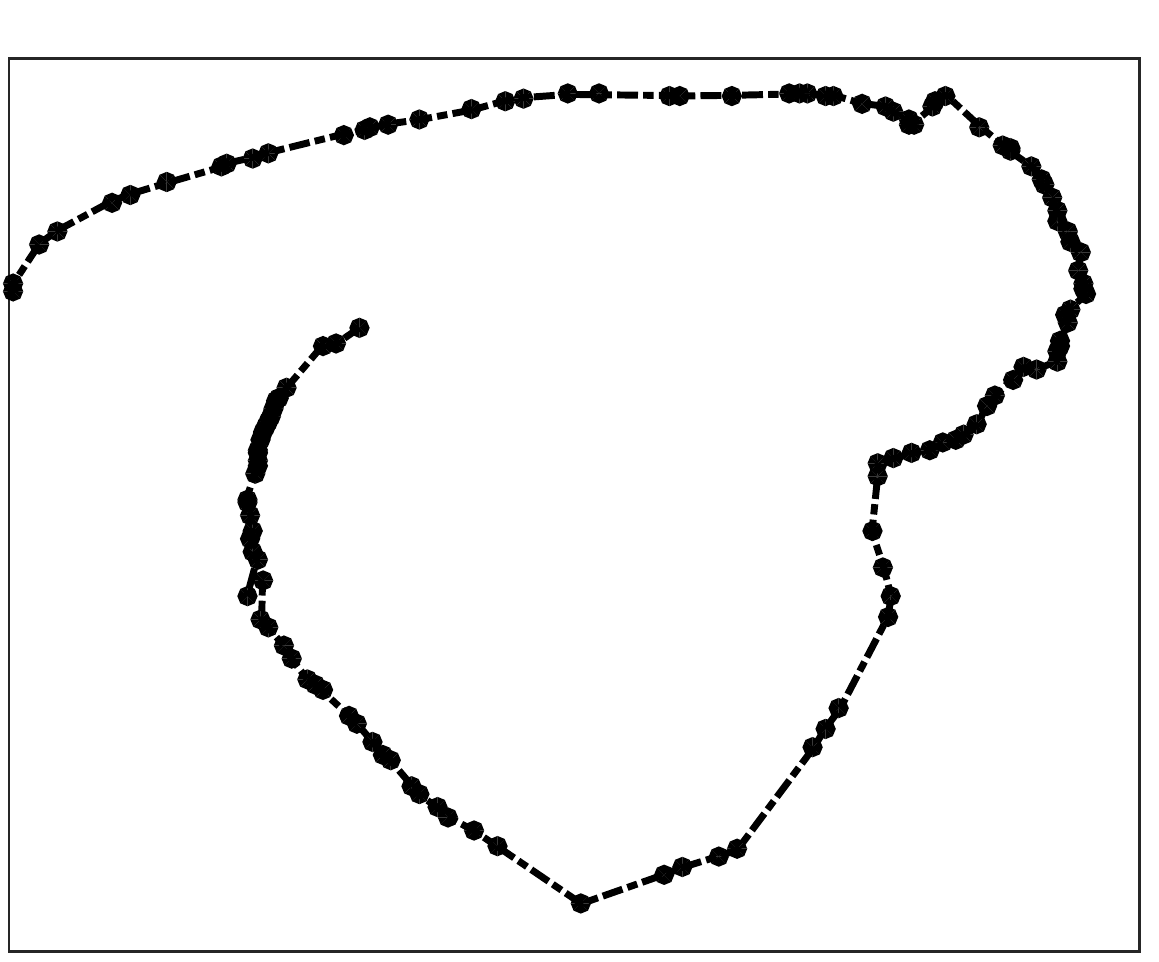} 
    \label{fig:out_trj_t}
  }%
  \subfigure[$\Delta{t}$ and Gain control]{
    \includegraphics[trim = 139 295 122 270, clip,width=0.3\columnwidth] {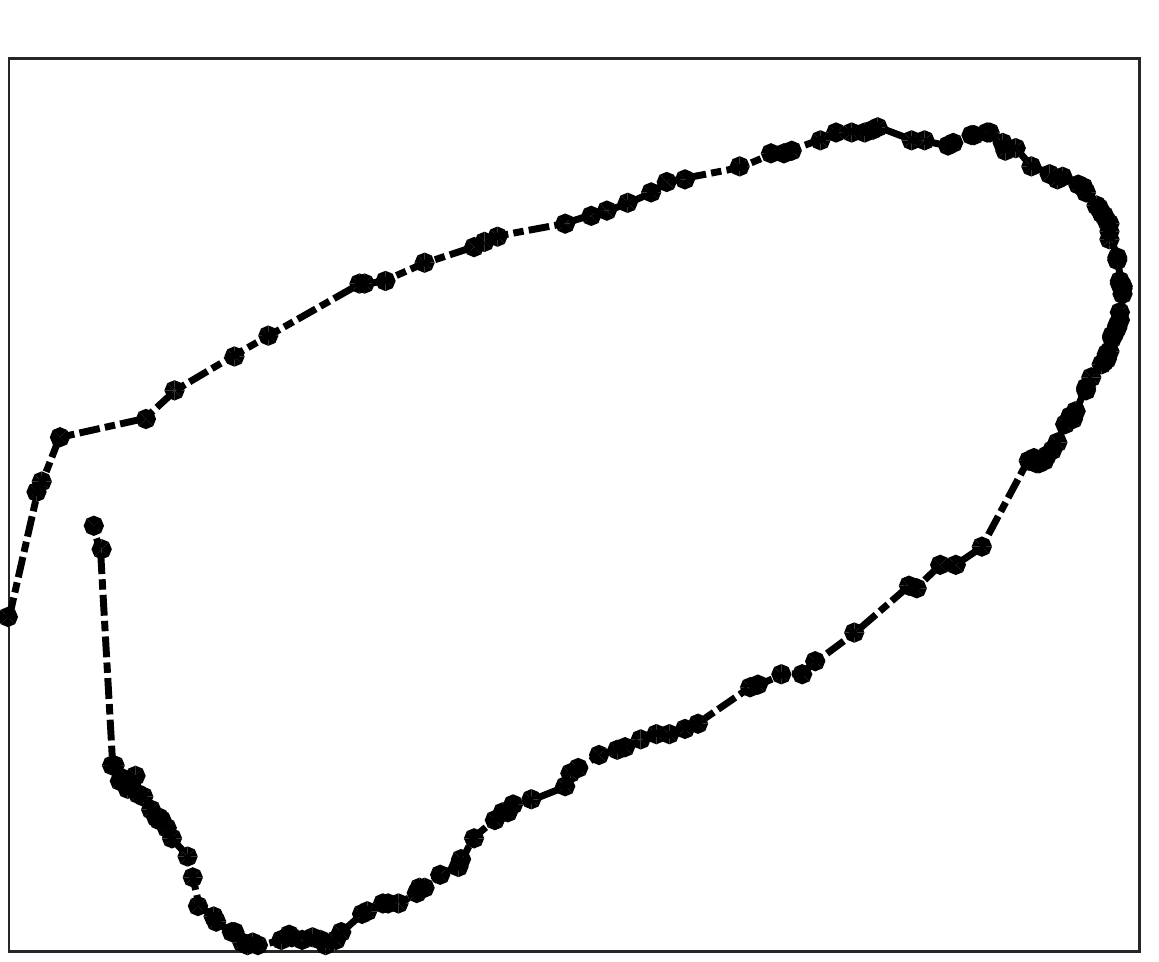} 
    \label{fig:out_trj_gt}
  }

  \caption{Trajectory of ORB SLAM in outdoor environment. The automatic
  attribute control (\texttt{AE}) caused noisy images from high gain and the ORB
  SLAM failed at the initialization phase, and thus excluded in the plot.}

  \label{fig:out_trj}
\end{figure}

We also verified the proposed method in a \ac{SLAM} framework, we piped the
acquired images in the ORB SLAM \cite{mur2015orb} (\figref{fig:out_trj}). In
both cases when using automatic exposure control and exposure time-only control,
tracking failure was caused by under-exposure. Refer to attached multimedia
\texttt{gcac.mp4} for more detailed tracking results. For the outdoor test,
automatic attribute control failed even at the initialization phase. Similar to
the indoor setting, handling gain and exposure time together
(\figref{fig:out_trj_gt}) outperformed exposure time only control
(\figref{fig:out_trj_t}), resulting in the more consistent trajectory.

\subsection{Discussion on Optimal Attribute and Computational Cost}
\label{sec:disc}

\begin{figure}[!h]
  \centering%
  \def\width{0.7\columnwidth}%
    \includegraphics[width=\width] {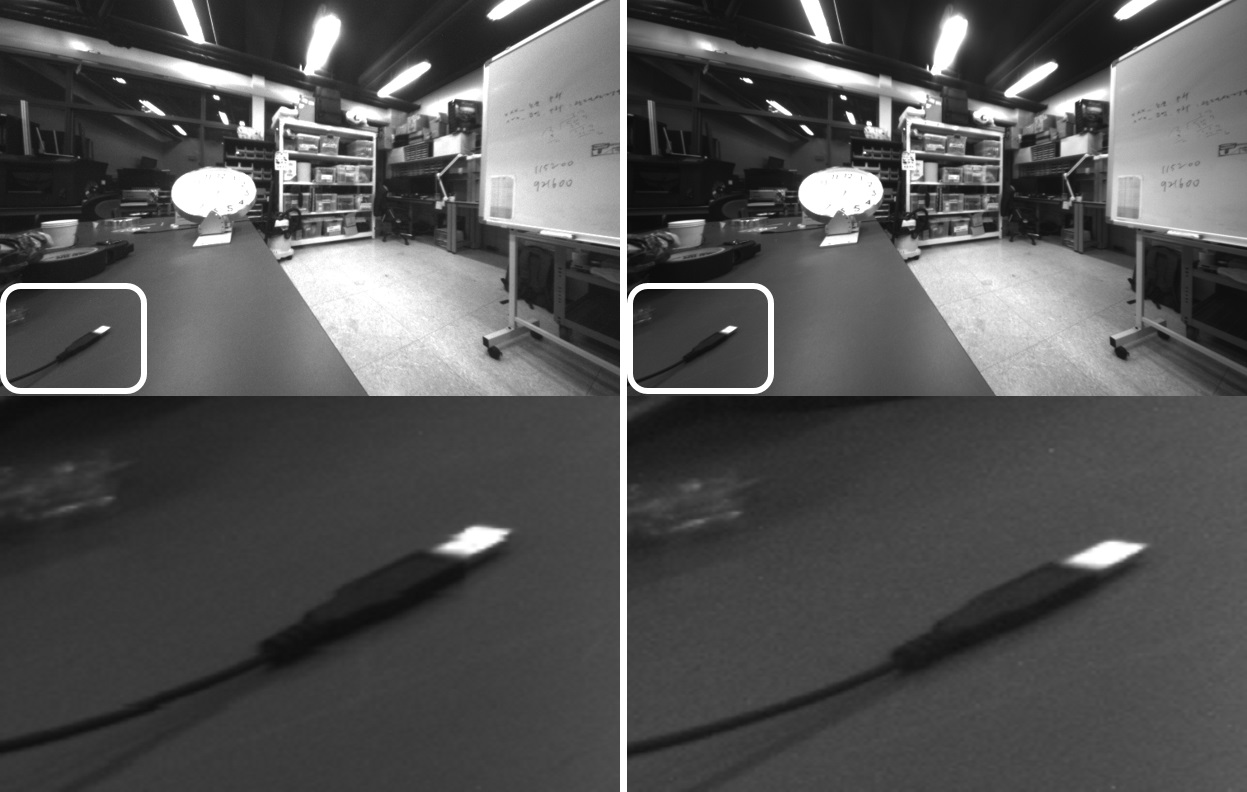}

  \caption{The left image is high exposure time and low gain,  the right image
  is low exposure and high gain image}

  \label{fig:disc}
\end{figure}

Selecting the optimal camera attribute is critical especially when light
conditions change in a dark environment. \figref{fig:disc} depicts two sample
images taken from different attribute pair. As can be seen in the zoomed view,
the noise level is substantially higher when high gain is assigned. This
indicates the importance of the joint control of the camera attributes.

Overall, our algorithm updates the exposure parameter to \unit{10}{Hz}. Using a
synthetic image for function evaluation substantially improve the performance
and ensure a real-time image stream. The detailed timing for each module is
summarized in the \tabref{tab:time}.

\begin{table}[!h]
  \centering
    \caption{Time required per each module}
    \label{tab:time}
    \begin{tabular}{c|cc|cc}
      Module   & Img. Synth. & Metric eval. & \multicolumn{2}{c}{Total Control}\\
               & [ms]        & [ms]         & [ms]    & [Hz]     \\
      \hline
      10 lux   & 0.254       & 13.780       & 101.278 & 9.87\\
      320 lux  & 0.247       & 12.768       & 94.215  & 10.61\\
      1000 lux & 0.246       & 13.242       & 93.364  & 10.71\\
    \end{tabular}
\end{table}

\section{conclusion}

This paper reported generic camera attribute control for both exposure time and
gain. The proposed control scheme simultaneously controls exposure time and gain
using fast function evaluation from \ac{CRF} based image synthesis. To the best
of our knowledge, the proposed method is the first unified and generic approach
to control exposure time and gain at the same time. We provided extensive
evaluations to discuss the relation between these two attributes in the
resulting images.



\bibliographystyle{IEEEtranN} 
\bibliography{string-short,references}

\end{document}